\documentclass[conference]{IEEEtran}
\IEEEoverridecommandlockouts
\usepackage{cite}
\usepackage{amsmath,amssymb,amsfonts}
\usepackage{algorithmic}
\usepackage{graphicx}
\usepackage{textcomp}
\usepackage{xcolor}
\usepackage{booktabs}

\usepackage{graphicx,xcolor}

\usepackage{array}
\usepackage{eqparbox}
\usepackage{multirow}
\usepackage{algorithm2e}
\usepackage{subcaption}
\usepackage{url}

\def\BibTeX{{\rm B\kern-.05em{\sc i\kern-.025em b}\kern-.08em
    T\kern-.1667em\lower.7ex\hbox{E}\kern-.125emX}}
\begin{document}

\title{Few-shot Class-Incremental Semantic Segmentation via Pseudo-Labeling and Knowledge Distillation\\
\thanks{This research was funded by Fujian NSF (2022J011112),
Research Project of Fashu Foundation (MFK23001),
and The Open Program of The Key Laboratory of Cognitive Computing and Intelligent Information Processing of Fujian Education Institutions, Wuyi University (KLCCIIP2020202).}

\author{\IEEEauthorblockN{Chengjia Jiang$^{1}$, Tao Wang$^{1,2,\ast}$, Sien Li$^{3}$, Jinyang Wang$^{3}$, Shirui Wang$^{1}$, Antonios Antoniou$^{4}$}
\IEEEauthorblockA{$^{1}$Fujian Provincial Key Laboratory of Information Processing and Intelligent Control, Minjiang University, Fuzhou, China. \\
$^{2}$The Key Laboratory of Cognitive Computing and Intelligent Information Processing of Fujian Education Institutions, \\Wuyi University, Wuyishan, China. \\
$^{3}$College of Computer and Data Science, Fuzhou University, Fuzhou, China.\\
$^{4}$Department of Computer Science and Engineering, European University Cyprus, Nicosia, Cyprus.}
$^{*}$Corresponding author: Tao Wang, E-mail: twang@mju.edu.cn.}
}

\maketitle

\begin{abstract}
We address the problem of learning new classes for semantic segmentation models from few examples, which is challenging because of the following two reasons.
   Firstly, it is difficult to learn from limited novel data to capture the underlying class distribution. Secondly, it is challenging to retain knowledge for existing classes and to avoid catastrophic forgetting.
   For learning from limited data, we propose a pseudo-labeling strategy to augment the few-shot training annotations in order to learn novel classes more effectively. Given only one or a few images labeled with the novel classes and a much larger set of unlabeled images, we transfer the knowledge from labeled images to unlabeled images with a coarse-to-fine pseudo-labeling approach in two steps. Specifically, we first match each labeled image to its nearest neighbors in the unlabeled image set at the scene level, in order to obtain images with a similar scene layout. This is followed by obtaining pseudo-labels within this neighborhood by applying classifiers learned on the few-shot annotations. In addition, we use knowledge distillation on both labeled and unlabeled data to retain knowledge on existing classes. We integrate the above steps into a single convolutional neural network with a unified learning objective.
   Extensive experiments on the Cityscapes and KITTI datasets validate the efficacy of the proposed approach in the self-driving domain. Code is available from~\url{https://github.com/ChasonJiang/FSCILSS}.
\end{abstract}

\begin{IEEEkeywords}
semantic segmentation, class-incremental learning, few-shot learning, knowledge distillation, pseudo-labeling
\end{IEEEkeywords}

\section{Introduction}

\begin{figure}[ht]
    \centering
    \includegraphics[trim={0.8cm 4cm 4.2cm 3cm},clip,width=0.5\textwidth]{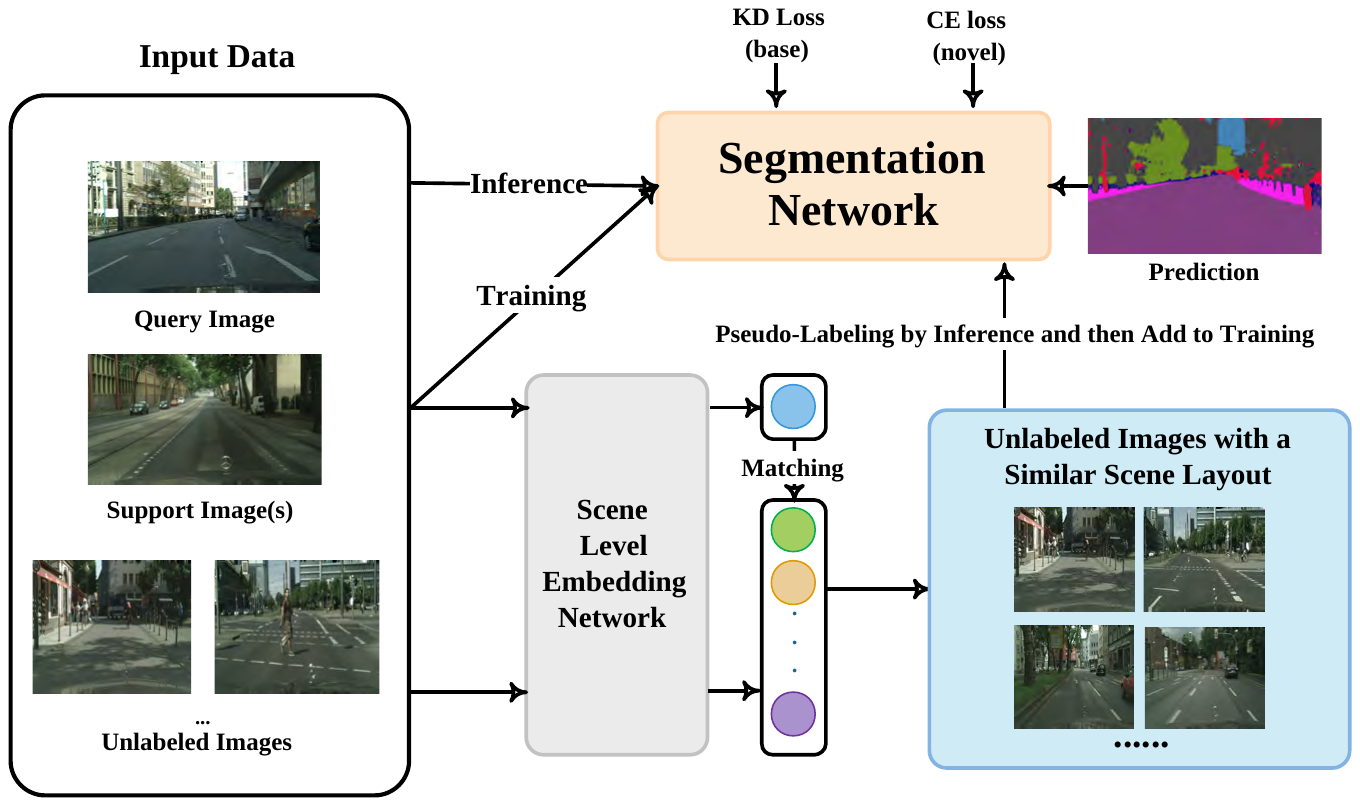}
    \caption{Overview of the proposed few-shot class-incremental semantic segmentation method in the self-driving domain. Given a query image, we use a scene-level embedding network to retrieve images with a similar scene layout from a large set of unlabeled images. These retrieved images are then labeled by inference via the segmentation network, and their pseudo-labels are, in turn, used to fine-tune the segmentation network for final prediction.}
    \label{fig:show}
\end{figure}

Semantic segmentation is a key perception component in advanced driver assistance and self-driving systems. In many practical applications, it would be desirable
to be able to add user-defined classes after the semantic segmentation models are initially trained and deployed.
Recently, Klingner et al.~\cite{klingner2020class} showed that we can perform class-incremental learning in this scenario with impressive performance.
Nevertheless, due to the time-consuming nature of the labeling process, it would be ideal for users to label only the novel classes in a small number of images and let the model perform learning in an incremental fashion. A similar problem in image classification was first presented in~\cite{tao2020few}, and the paradigm is referred to as few-shot class-incremental learning.
Naively learning with just a few examples under such setting, however, has severe limitations in terms of generalization abilities. In particular, standard data augmentation from a small number of examples would not suffice to capture the variations in the underlying class distribution.

A natural choice, therefore, would be to exploit unlabeled images in the user domain, which can be easily obtained. Given one or a few labeled query images, we can train an initial semantic segmentation model which are then used to generate pseudo-labels for unlabeled images.
However, since the initial model is trained with a small number of labeled images, it may not work well on a large unlabeled set of visually diverse images.
In this work, we propose a practical coarse-to-fine approach to pseudo-labeling. Specifically, for each query image we first search for its nearest neighbors at the scene level with scene-level features. This allows us to shortlist a set of unlabeled images with a similar scene layout and overall appearance. This is followed by applying the initial model on these visually similar images to obtain pixelwise pseudo-labels. We then apply a masked log loss function to learn novel classes with these additional labelings.
Another important aspect in the above learning paradigm is to retain knowledge on existing classes. In this work, we use knowledge distillation from a teacher model (i.e., the model previously learned from existing classes) on both labeled and unlabeled data, which has been proven effective~\cite{li2017learning,michieli2021knowledge} in class-incremental learning. Finally, we integrate the above learning steps into a single convolutional neural network with a unified learning objective. Recently, there have been a few attempts to address the class-incremental semantic segmentation problem under few-shot labeling constraints. For example, Cermelli et al.~\cite{cermelli2021prototype} proposed a prototype-based distillation loss to avoid overfitting and forgetting, and batch-renormalization to cope with non-\textit{i.i.d.} few-shot data. Shi et al.~\cite{shi2022incremental} proposed a method that uses a hyper-class embedding to store old knowledge and to adaptive update the category embedding of new classes. There are also a few recent methods proposed for instance segmentation~\cite{ganea2021incremental,nguyen2022ifs}. Unlike existing method, however, we make use of a scene-level embedding to obtain unlabeled images with a similar layout to the few-shot data and then integrate pseudo-labeling and knowledge distillation for effective incremental learning.

The contributions of our work are three-fold.
Firstly, we propose to explore the challenging problem of few-shot class-incremental semantic segmentation in the self-driving domain, and establish a performance baseline for standard knowledge distillation.
Secondly, we propose a simple yet effective strategy to obtain pseudo-labels for unlabeled user data.
Specifically, we first train an initial model with few-shot labeled data. We note that this initial model does not necessarily work well on all data. Therefore, in order to transfer the learned knowledge to unlabeled data with high accuracy, we match each labeled query image to its nearest neighbors in the unlabeled image dataset with a scene-level descriptor, and then apply the initial model within this neighborhood to obtain pseudo-labels. For each unlabeled image, we also mask out regions with less confident predictions so as to obtain more accurate pseudo-labels.
Finally, we integrate the above learning steps with knowledge distillation on both labeled and unlabeled data into a single convolutional neural network and demonstrate the superiority of the proposed method in terms of segmentation performance.



\section{Our Approach}
\label{sec:approach}

In this section, we formally introduce the few-shot class-incremental semantic segmentation problem and
our proposed approach in detail. Let us begin with the standard semantic segmentation problem first.
Given an input color image, semantic segmentation learns a mapping into a structured output with each pixel
labeled with a semantic class.
Mathematically, denote the input image as $\mathbf{x} \in \mathcal{R}^{H\times W\times 3}$, the semantic label set as
$\mathcal{C}=\{1,2,\dots,C\}$, our goal is to produce a structured semantic labeling $\mathbf{\hat{y}} \in \mathcal{C}^{H\times W}$.
During training, we are provided with a training dataset $\mathcal{D} = \{ (\mathbf{x}_i, \mathbf{y}_i) \}_{i=1}^{N} $,
where $\mathbf{x}_i$ and $\mathbf{y}_i$ are the $i$-th training image and its ground-truth semantic labeling, respectively.
We note that the size of the training dataset $N$ is usually a large number as opposed to the few-shot learning scenario
we discuss later in this section.

In this paper, we propose a few-shot class-incremental learning method for semantic segmentation. In our view, the key limitation that hinders this learning task is the lack of sufficient annotations for novel classes due to labor constraints on the user side. In many practical applications, it is not possible to label a large amount of data with the novel classes. We should, however, make a clear distinction between user data and user annotations. While annotations may be expensive, we could easily collect a large number of unlabeled images on the user side. This, along with other previous work in few-shot learning~\cite{su2020does,liu2020part}, inspire us to make use of unlabeled data for few-shot class-incremental learning.

More formally, at time step $t>1$, we are additionally given an unlabeled dataset $\mathcal{U}_t=\{ \tilde{\mathbf{x}}_i^t \}_{i=1}^{M_t}$, where $M_t$ could be large. Ideally, we want to be able to obtain pseudo-labels for all images in the above dataset, i.e., $\tilde{\mathcal{Y}}_t=\{ \tilde{\mathbf{y}}_i^t \}_{i=1}^{M_t}$. Since we have a few-shot training dataset $\mathcal{D}_t$, this could be conveniently done by applying the model trained on $\mathcal{D}_t$ to $\mathcal{U}_t$.
However, in the few-shot learning scenario, it is unlikely that the knowledge from $\mathcal{D}_t$ could generalize well to all data in $\mathcal{U}_t$. Therefore, for each labeled image $\mathbf{x}_i^t$ we retrieve its neighborhood $\mathcal{N}_i^t$ in $\mathcal{U}_t$. Together, the neighborhoods $ \mathcal{N}^t = \{ \mathcal{N}_1^t \cup \mathcal{N}_2^t \cup \cdots \mathcal{N}_{N_{t}}^t \}$ form a subset of $\mathcal{U}_t$ that we intend to obtain pseudo-labels on. 
A high-level overview of our method is presented in Figure~\ref{fig:show}.
In the following, we will describe the learning steps above in detail.

\subsection{Learning the Base Task}\label{AA}
Let us begin with learning for the base task $T_1$, which is a standard semantic segmentation problem.
Denote the input image and its corresponding ground-truth labeling as $\mathbf{x}$ and $\mathbf{y}$,
respectively. The semantic label set for $T_1$ is $\mathcal{C}_1=\{1,2, \cdots, C_1 \}.$ Note that both $\mathbf{x}$ and $\mathbf{y}$ can be indexed by a pixel location
$l \in \mathcal{L}=\{ 1,2,\cdots, H \cdot W\}$. In particular, we use $y_{(l,c)}$ to denote
the binary label for the $c$-th class at image location $l$.
During training, we are given a training dataset
$\mathcal{D}_1 = \{ (\mathbf{x}_i^1, \mathbf{y}_i^1) \}_{i=1}^{N_1}$. Our goal is to train a network $\mathcal{M}_1$ that maps an input image
to a structured semantic labeling, i.e., $\mathcal{M}_1: \mathcal{X} \mapsto \mathcal{Y}_1$ where
$\mathcal{X}$ and $\mathcal{Y}_1$ are the input image space and the output labeling space, respectively.
Denote the network output as $\mathbf{\hat{y}}$ and its class probability for the $c$-th class at
image location $l$ as $\hat{y}_{(l,c)}$, the cross-entropy loss function that we use to train $\mathcal{M}_1$ can be written as follows:

\begin{align}
    J_{\text{ce}}(\mathbf{y}, \mathbf{\hat{y}}) = - \frac{1}{|\mathcal{L}_{\mathcal{C}_{1}}|} \sum_{l \in \mathcal{L}_{\mathcal{C}_{1}}} \sum_{c \in \mathcal{C}_1} y_{(l,c)} \log ( \hat{y}_{(l,c)} )
\end{align}

\noindent where $\mathcal{L}_{\mathcal{C}_{1}} \subset \mathcal{L}$ is the set of \textit{labeled} pixel
locations, as we only consider classes defined by $\mathcal{C}_1$.

\subsection{Few-shot Class-Incremental Learning}

At time step $t>1$, we are given a class-incremental learning task $T_t$ and its corresponding few-shot training dataset $\mathcal{D}_t$. Similar to the base task, we train our network $\mathcal{M}_t$ with the cross-entropy loss function as follows:

\begin{align}
    J_{\text{ce}}(\mathbf{y}, \mathbf{\hat{y}}) = - \frac{1}{|\mathcal{L}_{\mathcal{C}_{t}}|} \sum_{l \in \mathcal{L}_{\mathcal{C}_{t}}} \sum_{c \in \mathcal{C}_t} y_{(l,c)} \log ( \hat{y}_{(l,c)} )
\end{align}

\noindent where $\mathcal{C}_t$ is the label set for the $t$-th task and $\mathcal{L}_{\mathcal{C}_{t}} \subset \mathcal{L}$ is the set of labeled pixel locations as defined by $\mathcal{C}_t$.

There are two key issues here: (1) how to learn our model with better quality when only limited labeled data is available, and (2) how to retain knowledge we learned from previous tasks with neither old data nor old labels. For the first issue, we propose a pseudo-labeling strategy on unlabeled data as described in the next subsection. For the second issue, we adopt a knowledge distillation~\cite{hinton2015distilling} approach with the following loss function:

\begin{align}
    J_{\text{kd}}(\bar{\mathbf{y}}, \mathbf{\hat{y}}) = - \frac{1}{|\mathcal{L} - \mathcal{L}_{\mathcal{C}_{t}}|} \sum_{l \not\in \mathcal{L}_{\mathcal{C}_{t}}} \sum_{c \in \mathcal{C}_{1:t-1}} \bar{y}_{(l,c)} \log ( \hat{y}_{(l,c)} )
\end{align}

\noindent where $\mathcal{C}_{1:t-1} = \{ \mathcal{C}_1 \cup \mathcal{C}_2 \cup \cdots \cup \mathcal{C}_{t-1} \}$ is the set of all previous classes. Particularly, we use $l \not\in \mathcal{L}_{\mathcal{C}_{t}}$ to denote the fact that the knowledge distillation loss is only computed on pixel locations that are not defined by $\mathcal{C}_t$. Since there are no ground-truth labelings available for these locations, we use $\bar{y}_{(l,c)}$ as a surrogate, which is the output class probability of the model from the previous time step $\mathcal{M}_{t-1}$.

\subsection{Pseudo-Labeling on Nearest Neighbors}
We now describe our pseudo-labeling strategy for class-incremental learning, which propagates the labels of the labeled images to their unlabeled neighbors. More specifically, at time step $t>1$, we are additionally given an unlabeled dataset $\mathcal{U}_t=\{ \tilde{\mathbf{x}}_i^t \}_{i=1}^{M_t}$.
Obtaining pseudo-labels for the entire unlabeled dataset is both time-consuming and inaccurate, so we adopt a practical two-step approach as follows.

\noindent \textbf{Building a scene-level neighborhood.} Our first step is to find the nearest neighbors of labeled images at the scene level. To this end, we use a model $\mathcal{M}_s$ to obtain a $Z$-dimensional scene-level feature embedding for each image in $\mathcal{D}_t$ and $\mathcal{U}_t$. In this work, we follow~\cite{wang2017efficient} and make use of the feature produced by the final average pooling layer of a ResNet-50~\cite{he2016deep} network pretrained on ImageNet~\cite{deng2009imagenet} for simplicity, and note that there could be better alternatives.
More specifically, let $\mathbf{F}^t \in \mathbb{R}^{Z \times N_t}$ and its
columns $\mathbf{f}_i^t, i= 1 \dots N_t$  denote features for the labeled images.
Similarly, let $\mathbf{G}^t \in \mathbb{R}^{Z \times M_t}$ and its columns
$\mathbf{g}_j^t, j = 1 \dots M_t$ denote features for the unlabeled images.
In practice, we also precompute a pairwise distance matrix $\mathbf{D}^t \in \mathcal{R}^{N_t \times M_t}$ whose $(i,j)$-th element $d_{ij}$ is the distance between $\mathbf{f}_i^t$ and $\mathbf{g}_j^t$,
i.e., $d_{ij} = \text{dist} (\mathbf{f}_i^t, \mathbf{g}_j^t)$. Here we use the cosine distance as the distance metric for $\text{dist}(\cdot, \cdot)$.
The above steps allow us to easily retrieve the nearest neighbors for each labeled image in $\mathcal{D}_t$, see Figure~\ref{fig:show} for an example. In this work, we choose $K$ nearest neighbors for each labeled image, and note that other neighborhood definitions may also be viable. At time step $t$, the neighborhood for $i$-th labeled image is denoted as $\mathcal{N}_i^t$.

\noindent \textbf{Pseudo-labeling by inference.}
Our second step is to obtain pseudo-labels for unlabeled images in $\mathcal{N}_t = \{ \mathcal{N}_1^t \cup \mathcal{N}_2^t \cup \cdots \mathcal{N}_{N_{t}}^t \}$. 
The basic idea is to transfer the labels to unlabeled images.
Therefore, we first use $\mathcal{D}_t$ to train $\mathcal{M}_t$, and then apply $\mathcal{M}_t$ to $\mathcal{N}^t$ to obtain pseudo-labels  $ \tilde{\mathbf{y}}_i^t, i = 1 \dots K \times N_t $.
We note that, in practice, in addition to using hard labels produced by the conventional argmax on the model prediction, we can also choose to use soft labels by retaining the probabilities in model prediction.
In this work, we use the former for its simplicity and note that we empirically found soft labels do not always produce superior results.
In this way, we obtain an additional training set $\mathcal{U}'_t=\{ (\tilde{\mathbf{x}}_i^t,\tilde{\mathbf{y}}_i^t) \}, \tilde{\mathbf{x}}_i^t \in \mathcal{N}^t, \mathcal{N}^t = \{ \mathcal{N}_1^t \cup \mathcal{N}_2^t \cup \cdots \mathcal{N}_{N_{t}}^t \}$ with pseudo-labels. Finally, we use $\mathcal{D}_t \cup \mathcal{U}'_t$ to retrain $\mathcal{M}_t$ to obtain the final model for the current time step.

\subsection{Model Learning}

At time step $t>1$, the overall learning objective of our method can be written as follows:

\begin{align}
    J(\mathbf{y}, \bar{\mathbf{y}}, \mathbf{\tilde{y}}, \mathbf{\hat{y}}) = \underbrace{J_{\text{ce}}(\mathbf{y}, \mathbf{\hat{y}})}_{\mathcal{D}_t~\text{(novel)}} + 
    \underbrace{J_{\text{ce}}(\mathbf{\tilde{y}}, \mathbf{\hat{y}})}_{\mathcal{U}'_t~\text{(novel)}} +
    \underbrace{J_{\text{kd}}(\bar{\mathbf{y}}, \mathbf{\hat{y}})}_{\mathcal{D}_t \cup \, \mathcal{U}'_t~\text{(base)}}
    \label{eqn:learn}
\end{align}

\noindent where the first and the third terms are used to obtain an initial $\mathcal{M}_t$ on $\mathcal{D}_t$ and, after the pseudo-labels are generated
with $\mathcal{M}_t$, all three terms are used to retrain $\mathcal{M}_t$ on $\mathcal{D}_t \cup \mathcal{U}'_t$ to obtain the final model for prediction.

\begin{table*}[t]
    \large
    \begin{center}
	\resizebox{0.75\linewidth}{!}{
        \begin{tabular}{c|c|ccccc}
    
    		\toprule
    		\multirow{2.5}{*}{\textbf{Methods}} & \multirow{2.5}{*}{\textbf{Stages}}&\multicolumn{5}{c}{Performance (mIoU)}  \\
    		
    		\cmidrule{3-7}
    		~ & ~ & $T_1$ & $T_2$ & $T_{1\cup2}$ & $T_3$ & $T_{1\cup2\cup3}$\\
            \midrule
            \multicolumn{7}{l}{\textbf{1-shot class-incremental learning}}\\
    		\midrule
    		
    		FT+KD & $T_1,T_2$ & $79.7\pm{1.30}$ & $\mathbf{16.2\pm{0.44}}$ & $37.6\pm{0.83}$ & - & - \\
    		
    		FT+KD+PL  & $T_1,T_2$ & $\mathbf{80.2\pm{0.70}}$ & $15.8\pm{0.53}$ & $\mathbf{39.4\pm{0.70}}$ & - & - \\
    		\midrule
    		FT+KD  & $T_1,T_2,T_3$ & $69.7\pm{1.10}$ & $\mathbf{15.8\pm{0.35}}$ & $31.1\pm{0.53}$ & $11.8\pm{0.88}$ & $20.5\pm{0.44}$ \\
    
      	FT+KD+PL  & $T_1,T_2,T_3$ & $\mathbf{73.1\pm{0.75}}$& 
            $14.2\pm{0.01}$ & $\mathbf{36.7\pm{0.35}}$ & $\mathbf{12.9\pm{0.79}}$ & $\mathbf{25.0\pm{0.31}}$ \\
    
            \midrule
            \multicolumn{7}{l}{\textbf{5-shot class-incremental learning}}\\
    		\midrule
            FT+KD & $T_1,T_2$ & $\mathbf{81.8\pm{0.70}}$ & $24.5\pm{1.50}$ & $45.0\pm{0.83}$  & - & - \\
            FT+KD+PL & $T_1,T_2$ & $81.6\pm{0.44}$ & $\mathbf{29.8\pm{1.30}}$ & $\mathbf{47.4\pm{0.66}}$  & - & - \\
            \midrule
            FT+KD & $T_1,T_2,T_3$ & $75.1\pm{0.66}$ & $21.9\pm{0.70}$ & $41.3\pm{0.53}$ & $27.1\pm{0.96}$ & $32.6\pm{0.80}$ \\
            FT+KD+PL & $T_1,T_2,T_3$ & $\mathbf{77.2\pm{0.57}}$ & $\mathbf{25.2\pm{0.22}}$ & $\mathbf{43.6\pm{0.26}}$ & $\mathbf{33.2\pm{0.75}}$ & $\mathbf{35.9\pm{0.35}}$ \\
    		\bottomrule
    	\end{tabular}
    }
     \end{center}
    \caption{Quantitative performance (mIou) of few-shot class-incremental semantic segmentation on the \textbf{Cityscapes} dataset. Here, $T_1$ is the base
    task, and $T_2$ and $T_3$ are the two sequential incremental learning tasks. FT, KD and PL refer to Fine-Tuning, Knowledge Distillation and Pseudo-Labeling, respectively.
    Compared to the baseline (FT+KD), when our pseudo-labeling strategy is added (FT+KD+PL), the performance is significantly improved under most settings.}
    \label{tab:cityscapes}
\end{table*}

\begin{table*}[t]
    \large
    \begin{center}
	\resizebox{0.5\linewidth}{!}{
        
        \begin{tabular}{c|c|ccc}
    
    		\toprule
    		\multirow{2.5}{*}{\textbf{Methods}} & \multirow{2.5}{*}{\textbf{Stages}}&\multicolumn{3}{c}{Performance (mIoU)}  \\
    		
    		\cmidrule{3-5}
    		~ & ~ & $T_1$ & $T_2$ & $T_{1 \cup 2}$ \\ 
            \midrule
            \multicolumn{5}{c}{\textbf{1-shot class-incremental learning}}\\
    		\midrule
    		
            FT+KD & $T_1,T_2$ & $47.9\pm{3.80}$ & $\mathbf{61.7\pm{6.20}}$ & $40.7\pm{2.80}$ \\
            FT+KD+PL & $T_1,T_2$ & $\mathbf{51.7\pm{2.40}}$ & $56.3\pm{9.20}$ & $\mathbf{42.5\pm{2.40}}$ \\

            \midrule
            \multicolumn{5}{c}{\textbf{5-shot class-incremental learning}}\\
    		\midrule
            FT+KD  & $T_1,T_2$ & $60.9\pm{1.50}$ & $77.2\pm{4.10}$ & $56.7\pm{2.50}$ \\
            FT+KD+PL & $T_1,T_2$ & $\mathbf{61.7\pm{0.62}}$ & $\mathbf{81.7\pm{4.60}}$ & $\mathbf{57.5\pm{2.50}}$ \\
    		\bottomrule
    	\end{tabular}
  }
  \end{center}
  \caption{Quantitative performance (mIoU) of few-shot class-incremental semantic segmentation on the \textbf{KITTI} dataset. Here, $T_1$ is the base task and $T_2$ is the incremental learning task. FT, KD and PL refer to Fine-Tuning, Knowledge Distillation and Pseudo-Labeling, respectively.
    Compared to the baseline (FT+KD), when our pseudo-labeling strategy is added (FT+KD+PL), the performance is significantly improved under most settings.}
    \label{tab:kitti}
\end{table*}

\section{Experimental Evaluation}
\label{sec:experiments}


In this section, let us describe the experimental evaluation details on Cityscapes~\cite{cordts2015cityscapes} and KITTI~\cite{abu2018augmented} datasets using our method. We begin by reporting the details regarding the datasets being used and our experimental setup in Section~\ref{sec:exp:datasets}, and the present the quantitative and qualitative results we obtained in Section~\ref{sec:exp:res}.
\subsection{Datasets and Experimental Settings}
\label{sec:exp:datasets}
In this work, we present the experimental results we obtained on two publicly available datasets, Cityscapes and KITTI, in order to evaluate
the performance of the proposed method in the autonomous driving domain. In all our experiments, we follow the experimental evaluation
settings from~\cite{klingner2020class}.

For the Cityscapes dataset, we follow~\cite{klingner2020class} and divide it into three subsets for one base learning task and two incremental learning tasks,
and leave out an additional validation set for performance evaluation.
Specifically, the subset for the base task include images from Aachen, Dusseldorf, Hannover and Strasbourg, and the subset for the first incremental task include
images from Bochum, Hamburg, Jena, Monchengladbach, Ulm, Weimar and Tubingen, and the subset for the second incremental learning task include
images from Bremen, Cologne, Stuttgart, Darmstadt, Krefeld and Zurich. Finally, we use images from Frankfurt, Lindau and Munster for validation.

For the KITTI dataset, due to its relatively small size, we use the base model trained on Cityscapes and consider only one incremental task with KITTI.
We note that this setting presents a unique challenge for the model trained on Cityscapes to adapt to image scenes in KITTI.
Specifically, we use the first 100 images for training the incremental learning task, and the other 100 images for validation.

In terms of the classes used for the base and incremental tasks, we also follow~\cite{klingner2020class} and use
road, sidewalk, vegetation, terrain and sky for the base task on both datasets. For Cityscapes, the classes for
the first incremental learning task are building, wall, fence, pole, traffic light and traffic sign, and the
classes for the second incremental learning task are person, rider, car, truck, bus, train, motorcycle and bicycle.
Again, due to the limited size of the KITTI datasets, we consider only two classes, car and building, for its incremental learning task.

For the class incremental tasks, we present results obtained with 1-shot and 5-shot respectively for the categories included in each increment.
Importantly, the results may vary as we only use a very small number of labeled images in each learning task, so we carry out 20 independent experiments
on the Cityscapes dataset and report the mean and the 95\% confidence interval. We carry out 10 independent experiments on the KITTI dataset.
In terms of the neighborhood size $K$, we empirically choose $10$ for both 1-shot and 5-shot learning on Cityscapes, and $1$ and $5$ respectively for 1-shot and 5-shot learning
on KITTI. In all our experiments, we follow standard practice in semantic segmentation and use mean IoU (mIoU)~\cite{lin2014microsoft} as the performance evaluation metric.

\begin{figure*}[h!]
    \centering
    \begin{minipage}{0.24\linewidth}
        \centering
        \caption*{Image}
        \includegraphics[width=\linewidth]{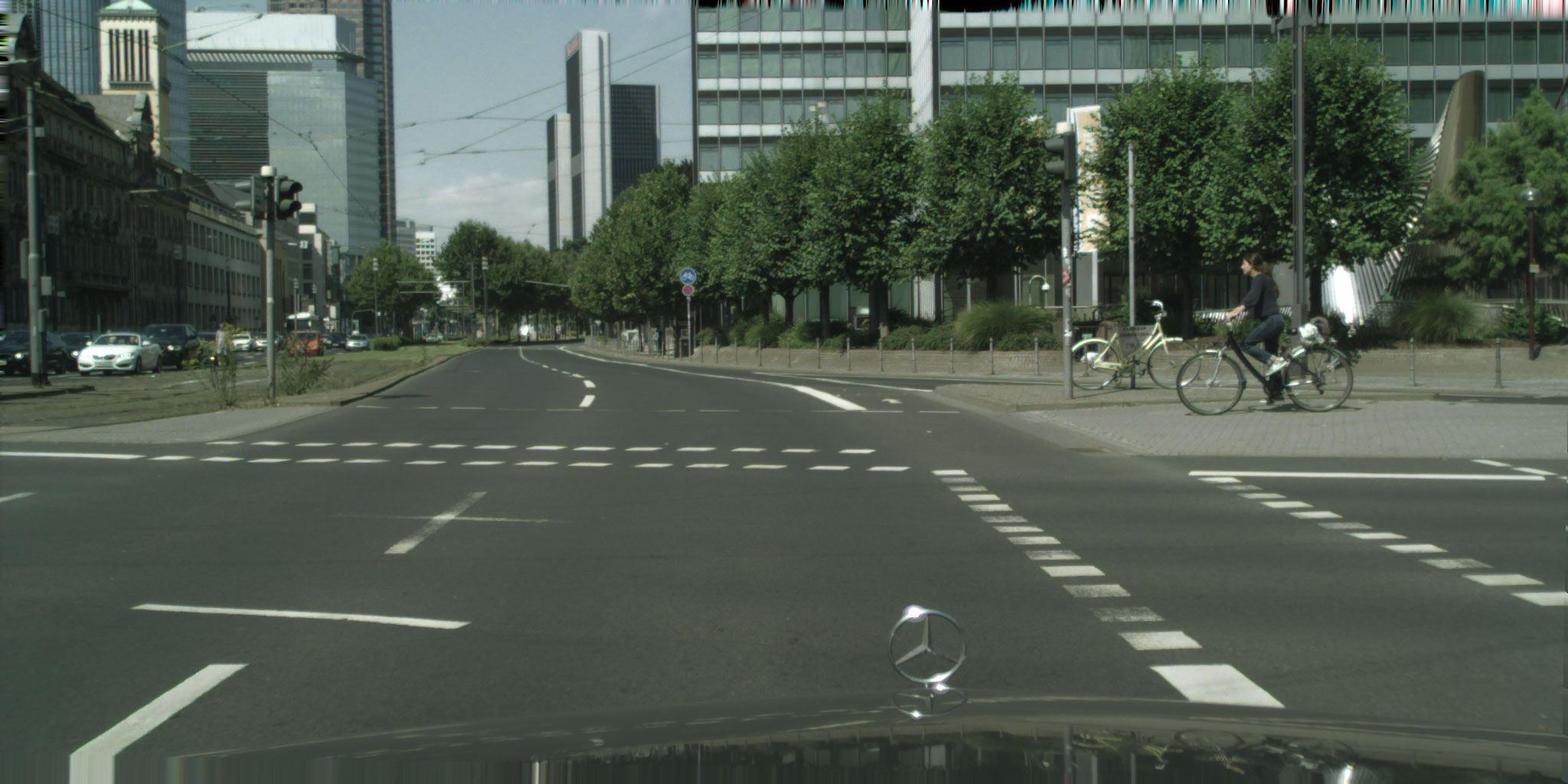}\vspace{0.1cm}
        \includegraphics[width=\linewidth]{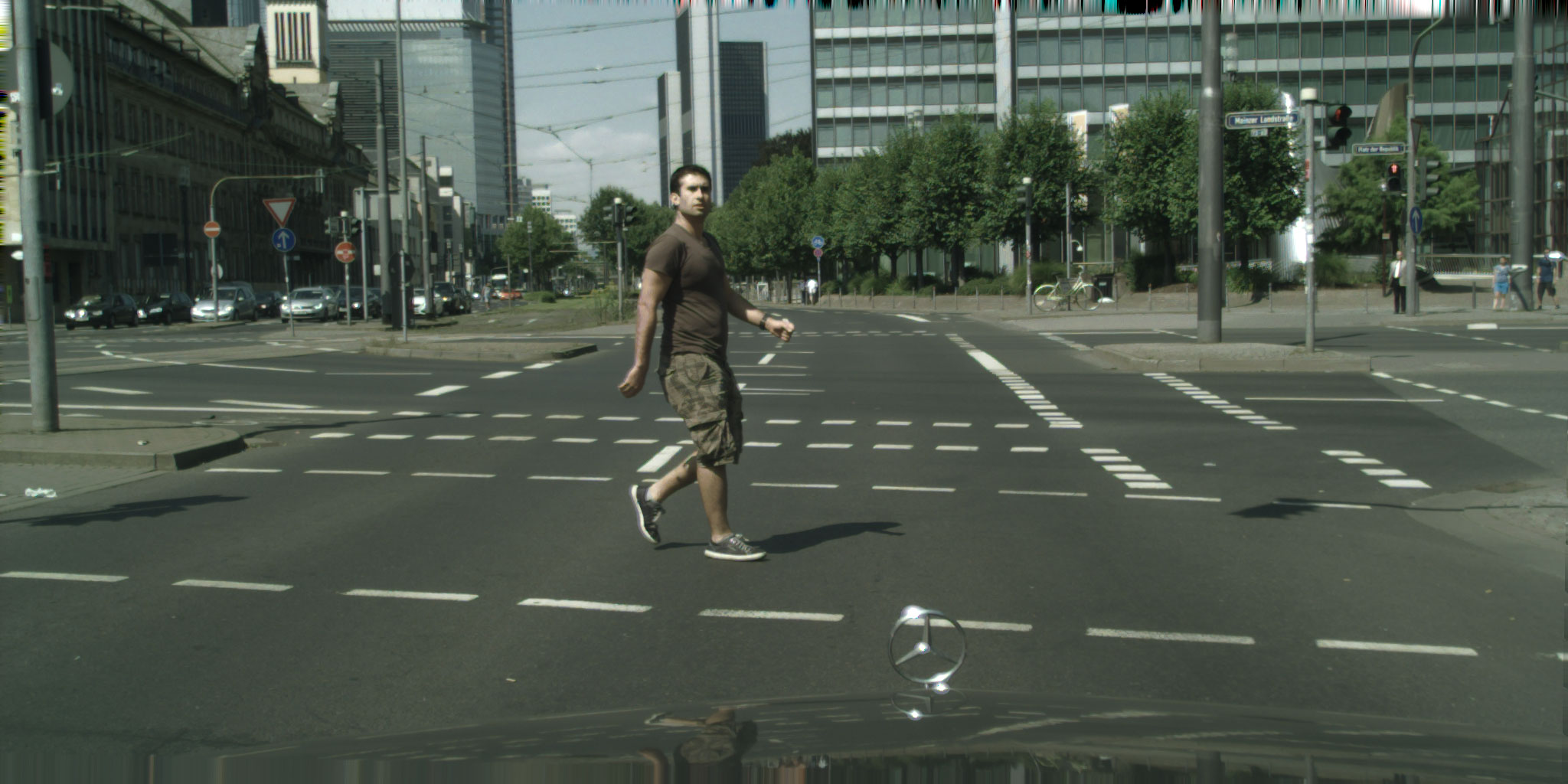}\vspace{0.1cm}
        \includegraphics[width=\linewidth]{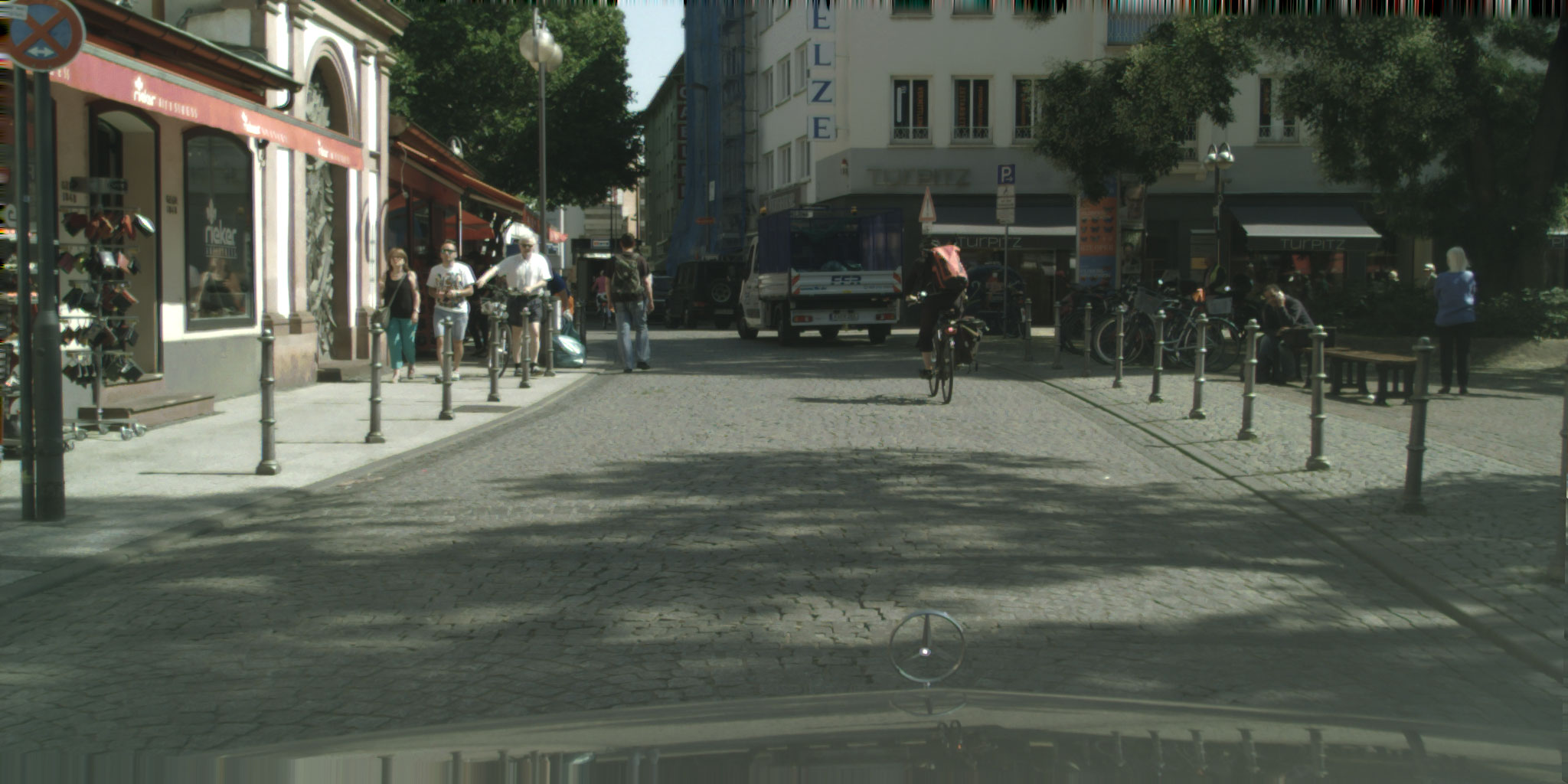}\vspace{0.1cm}
        \includegraphics[width=\linewidth]{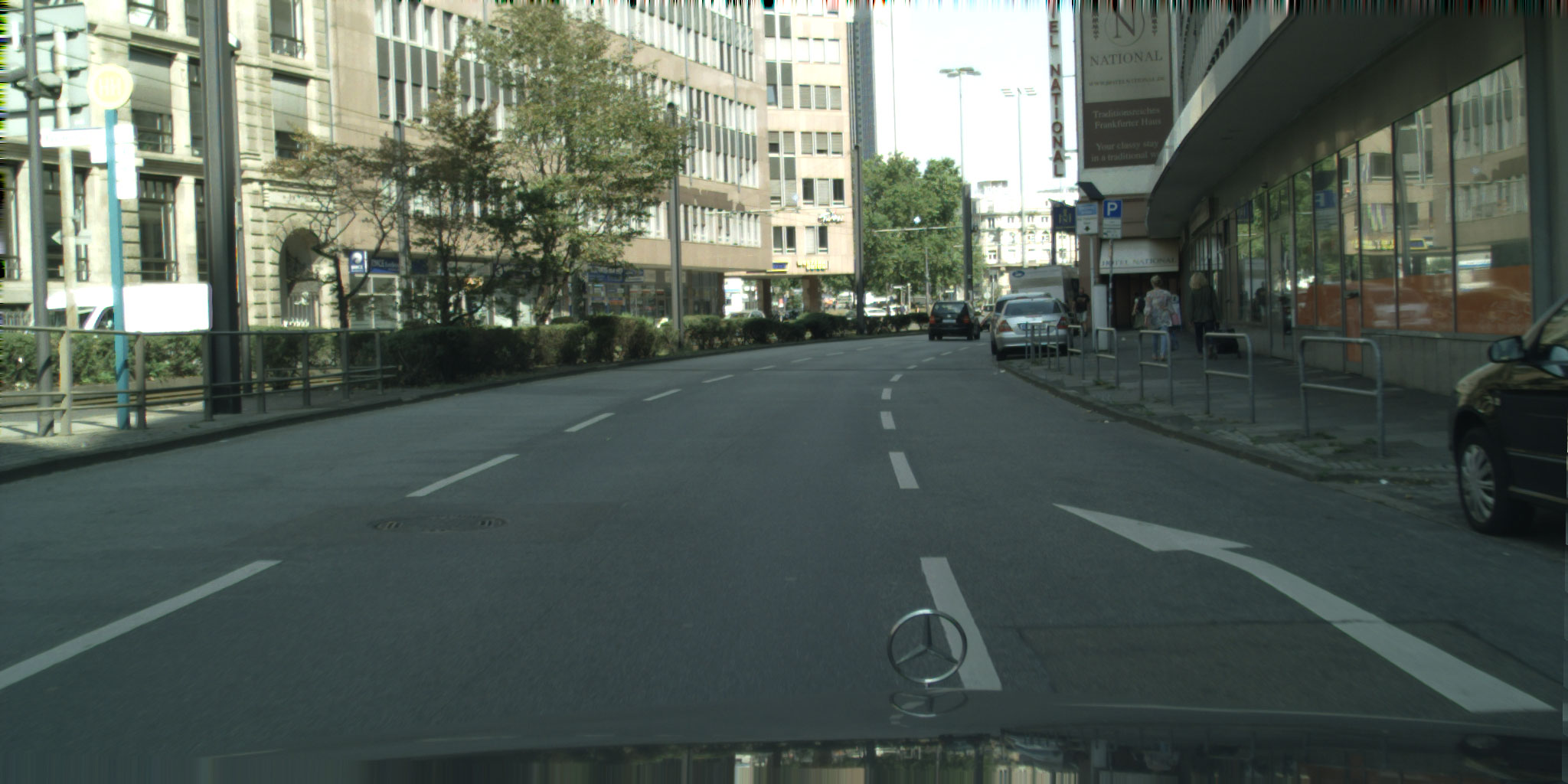}\vspace{0.1cm}
        \includegraphics[width=\linewidth]{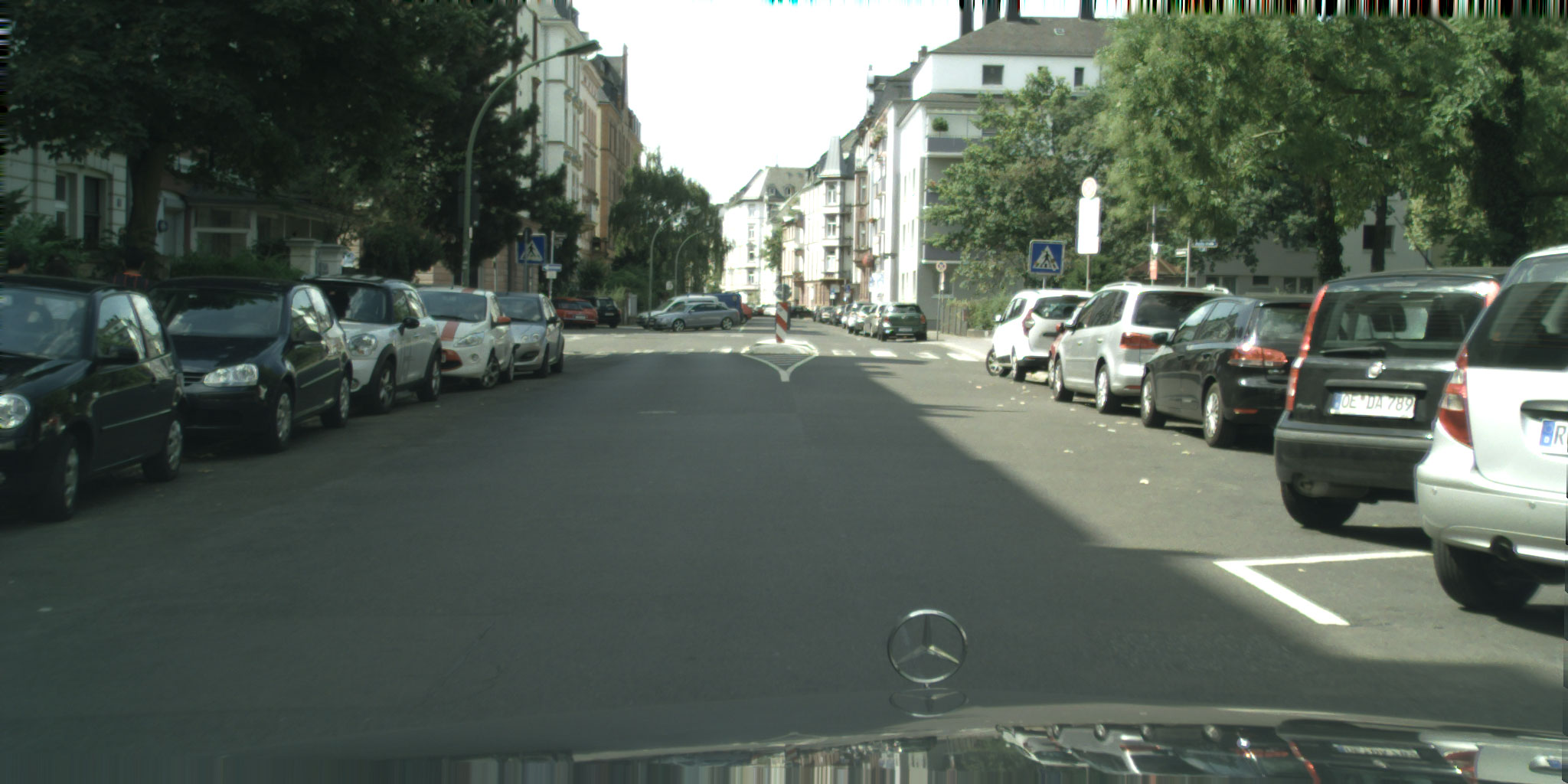}\vspace{0.1cm}
        \includegraphics[width=\linewidth]{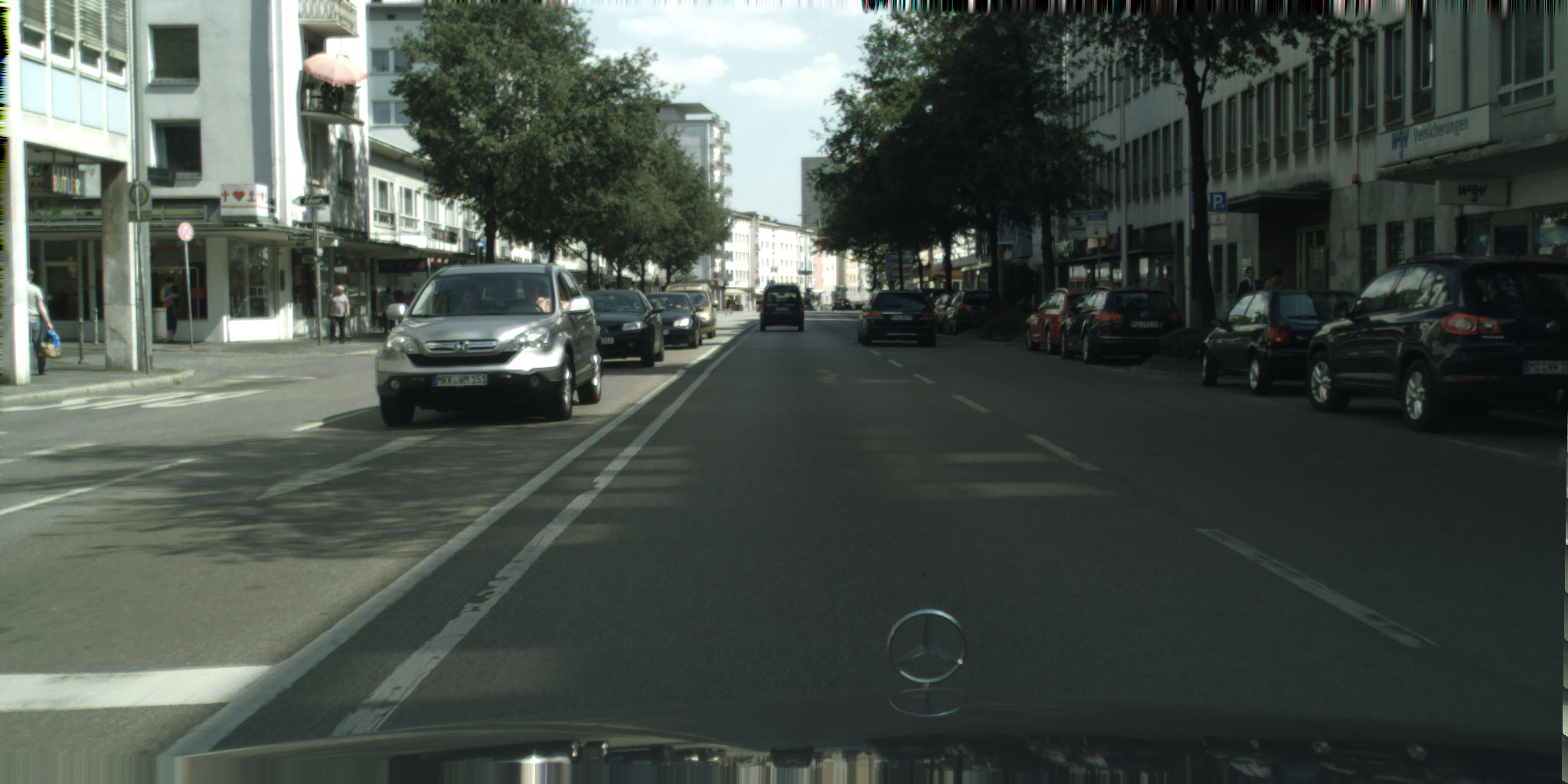}\vspace{0.1cm}
        \includegraphics[width=\linewidth]{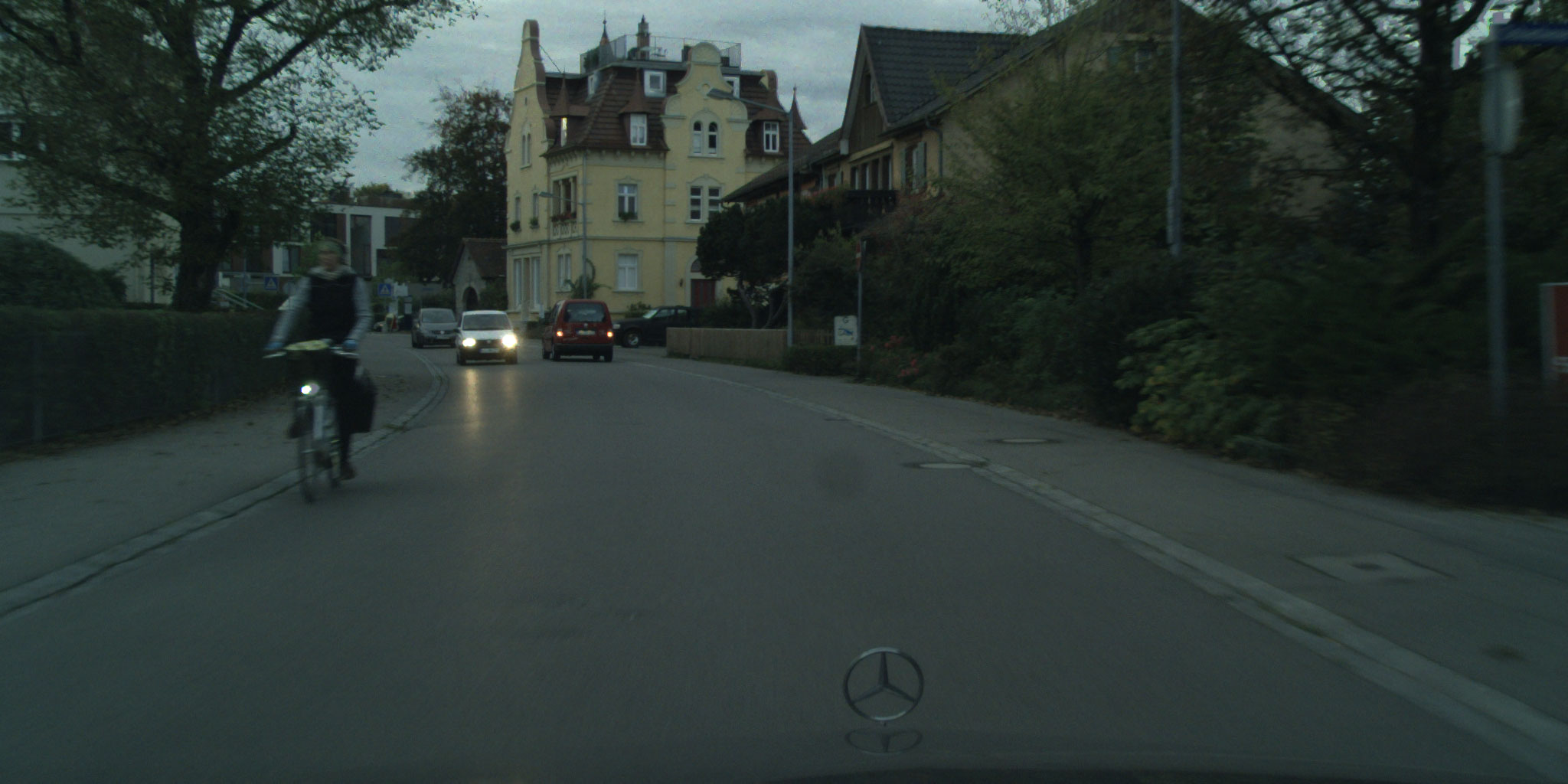}\vspace{0.1cm}
        \includegraphics[width=\linewidth]{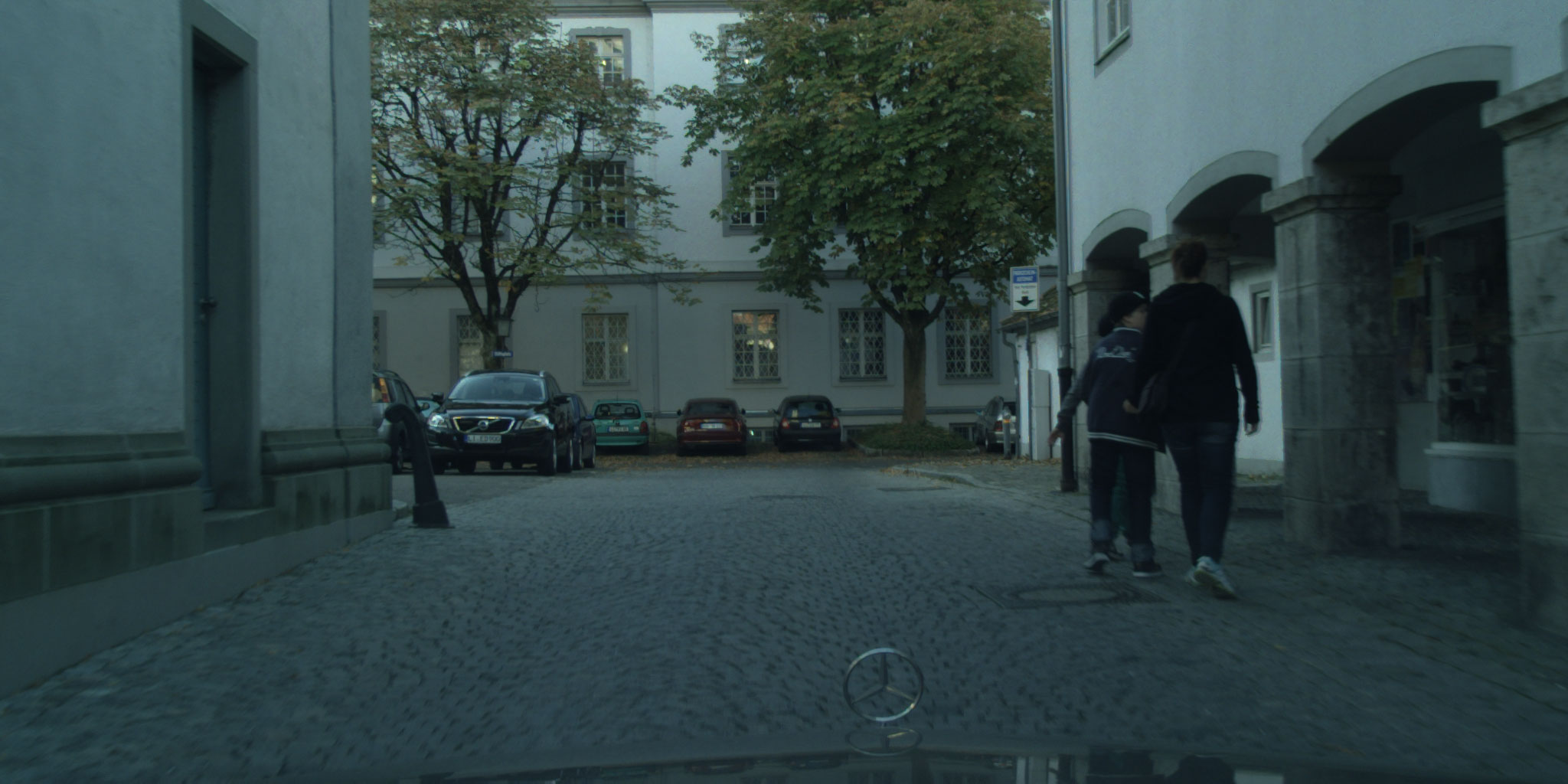}
    \end{minipage}\hfill
    \begin{minipage}{0.24\linewidth}
        \centering
        \caption*{GT}
        \includegraphics[width=\linewidth]{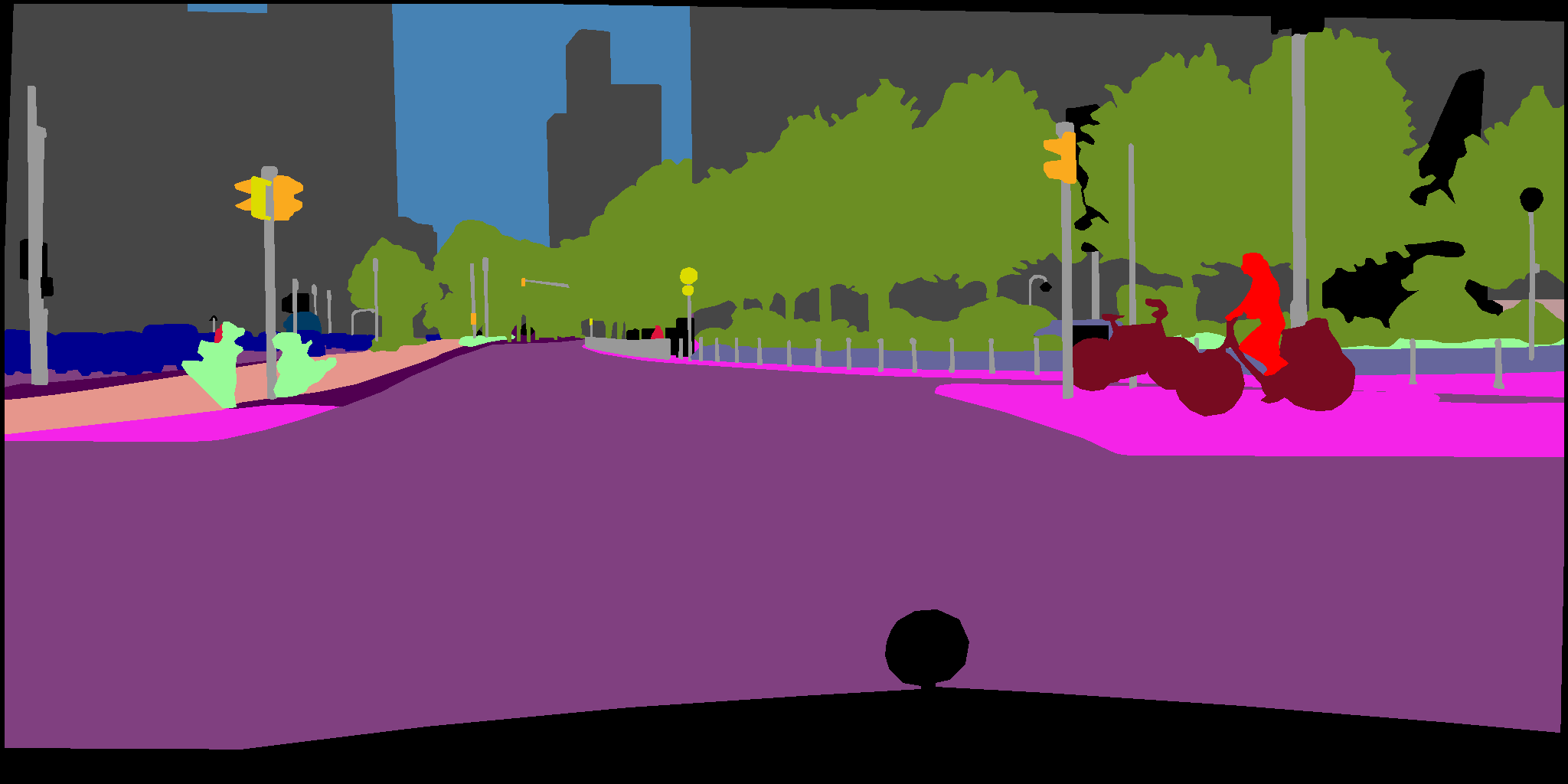}\vspace{0.1cm}
        \includegraphics[width=\linewidth]{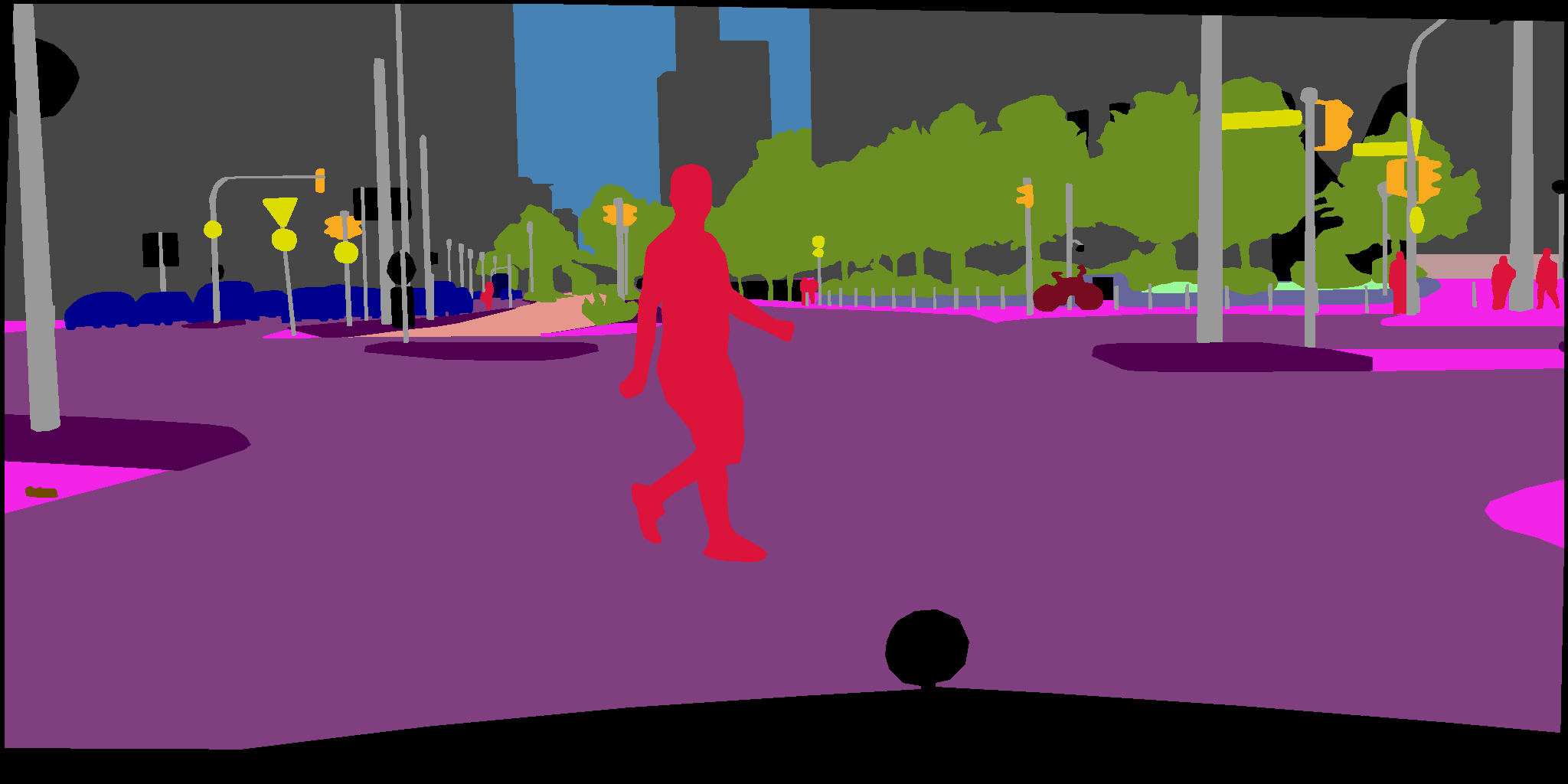}\vspace{0.1cm}
        \includegraphics[width=\linewidth]{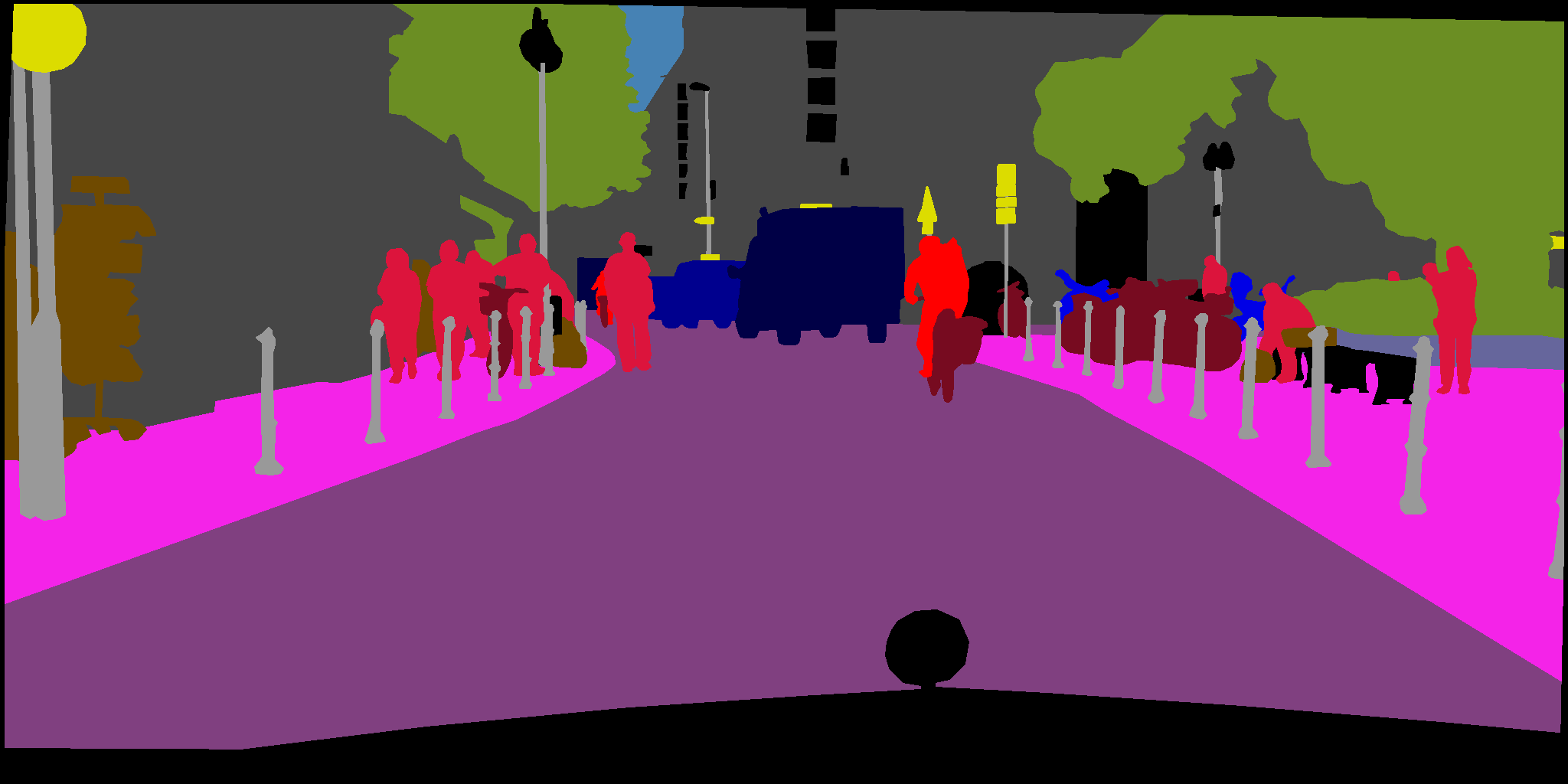}\vspace{0.1cm}
        \includegraphics[width=\linewidth]{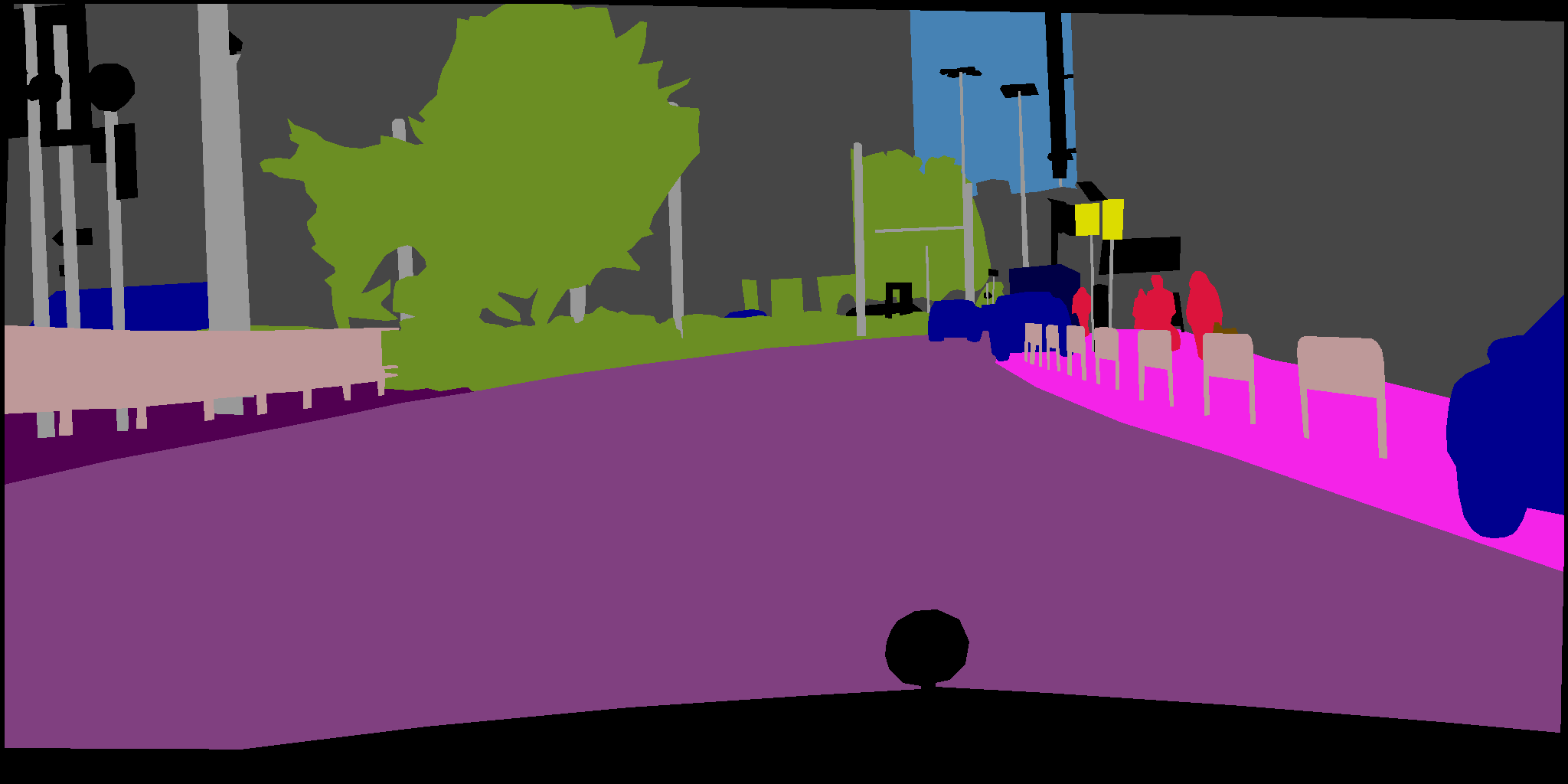}\vspace{0.1cm}
        \includegraphics[width=\linewidth]{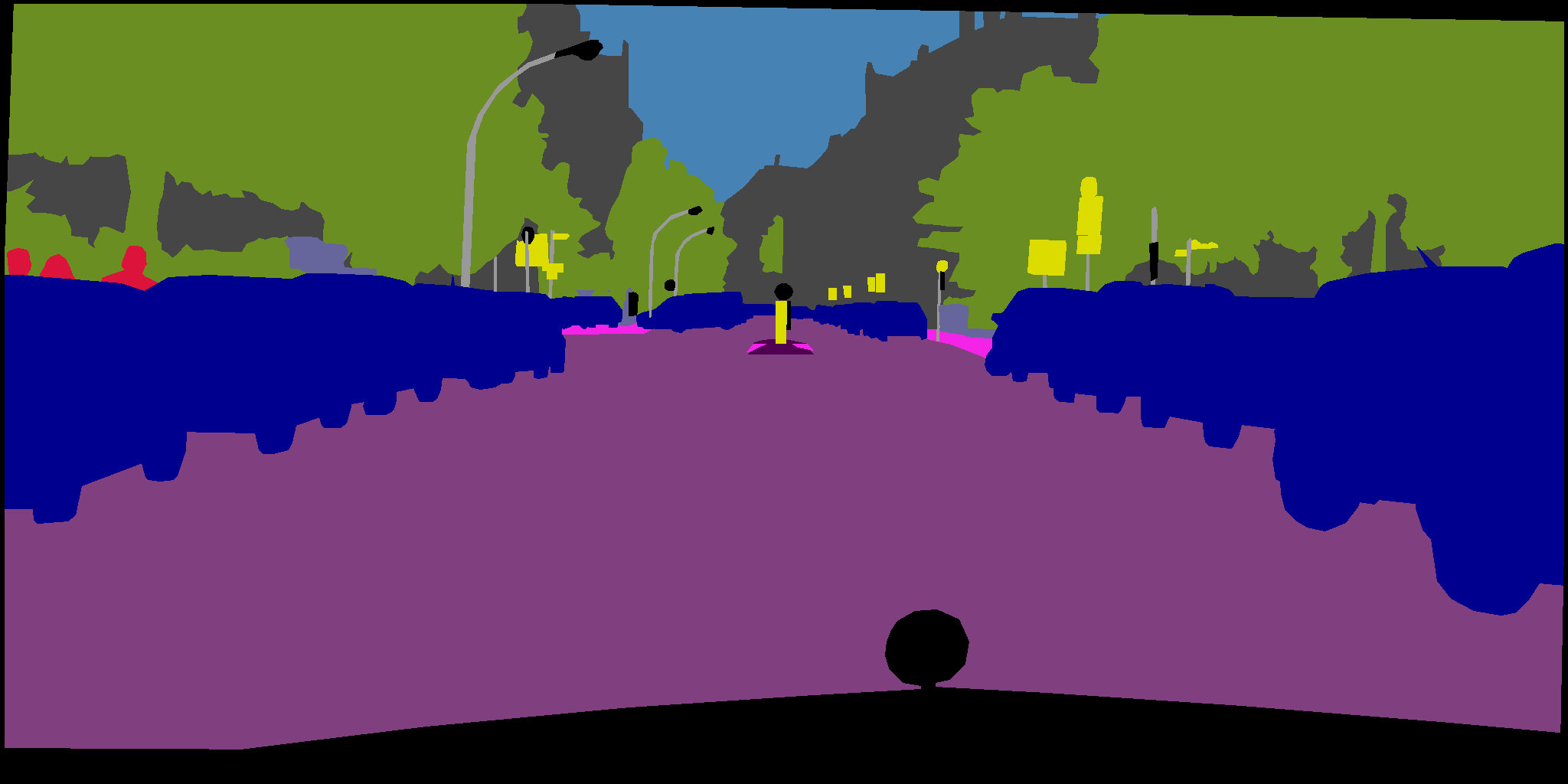}\vspace{0.1cm}
        \includegraphics[width=\linewidth]{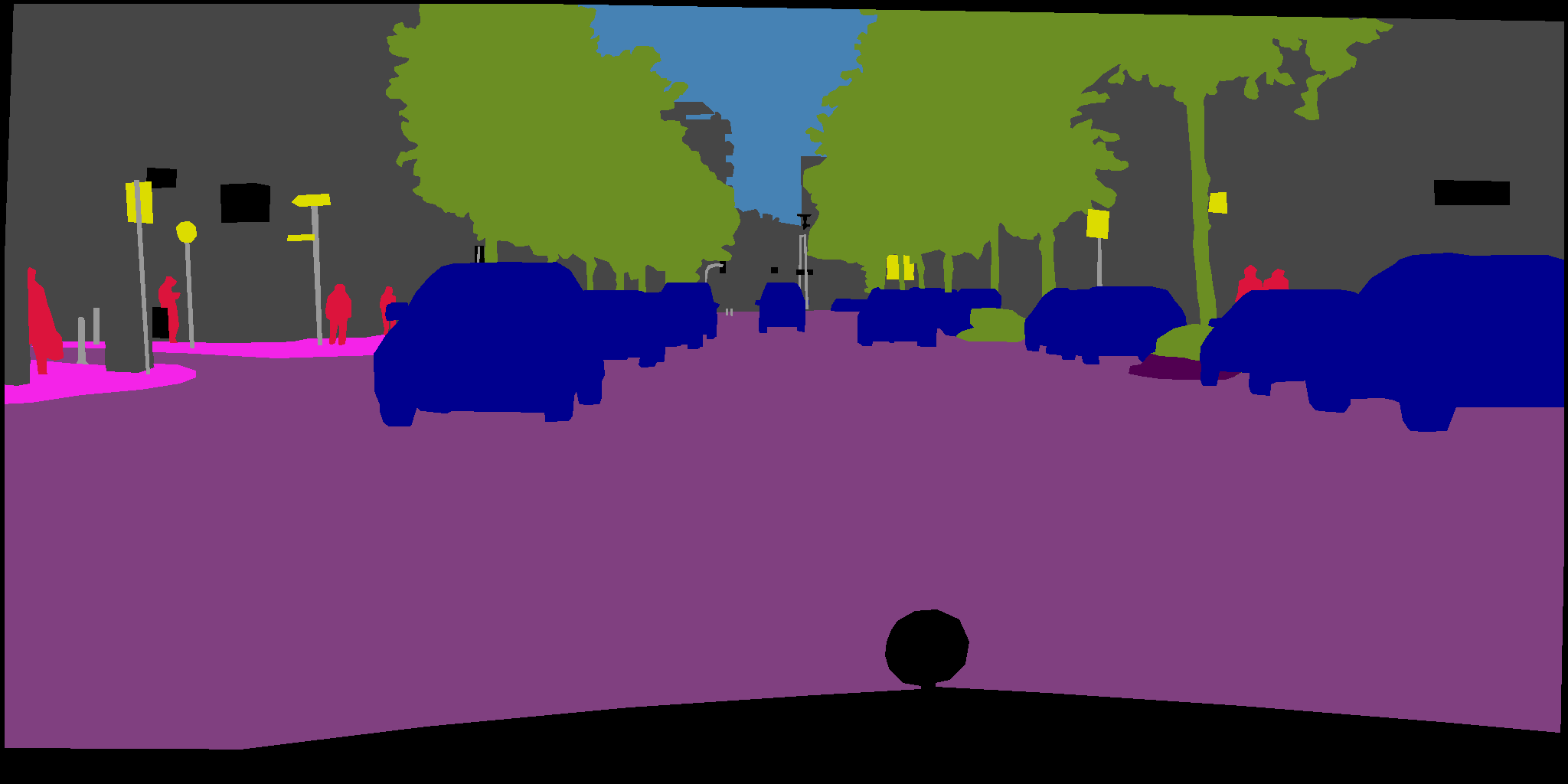}\vspace{0.1cm}
        \includegraphics[width=\linewidth]{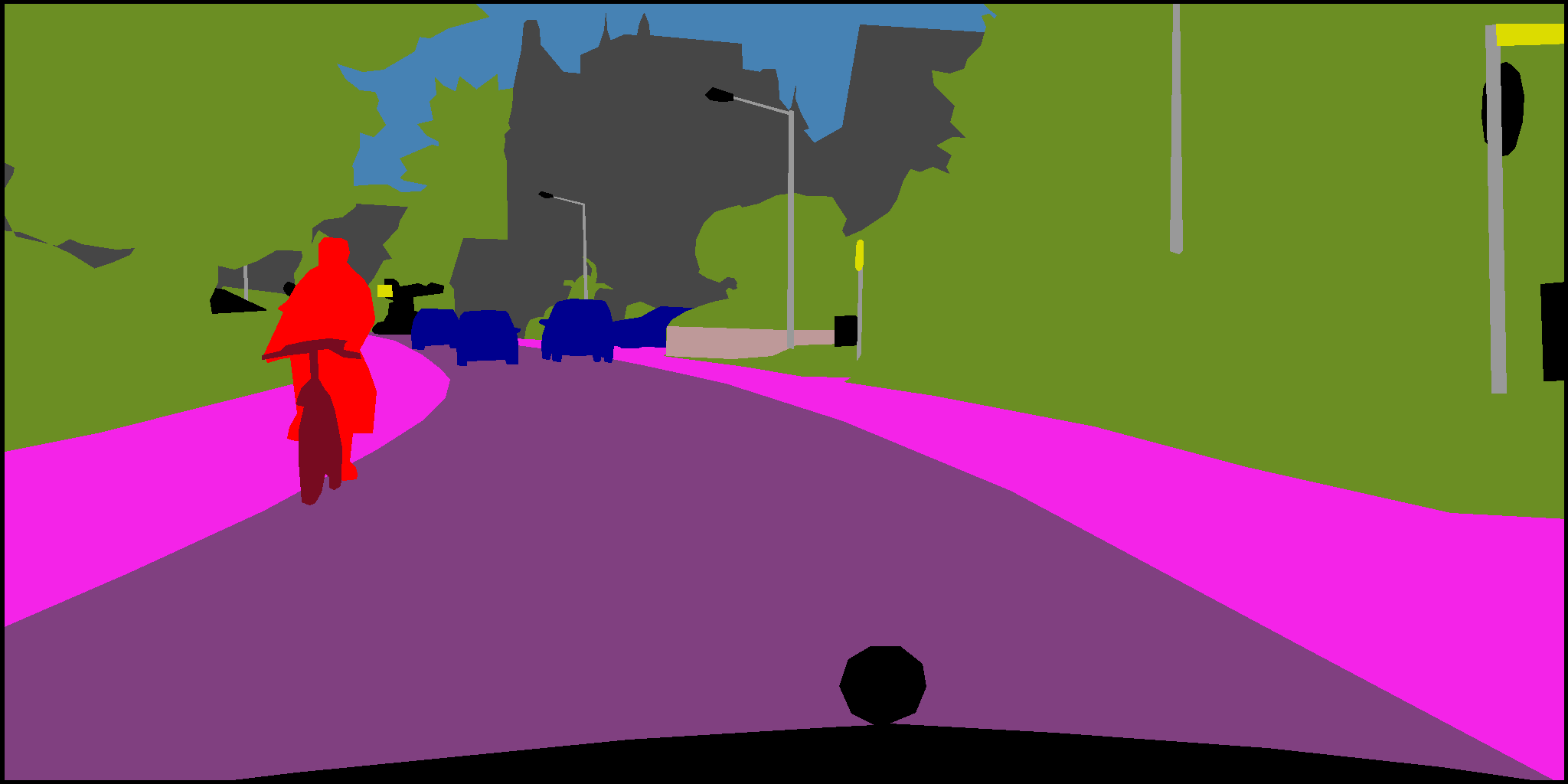}\vspace{0.1cm}
        \includegraphics[width=\linewidth]{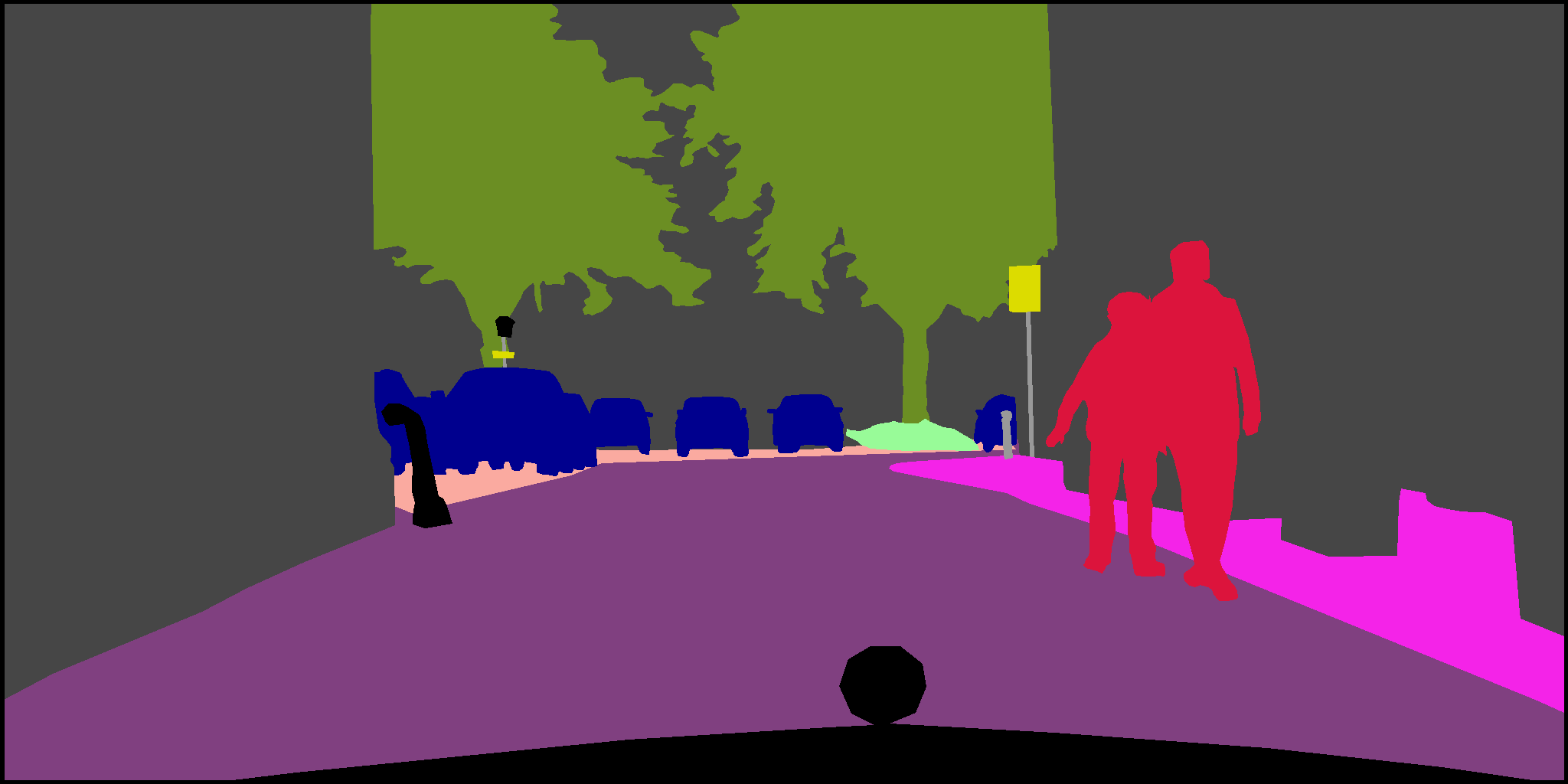}
    \end{minipage}\hfill
    \begin{minipage}{0.24\linewidth}
        \centering
        \caption*{Baseline}
        \includegraphics[width=\linewidth]{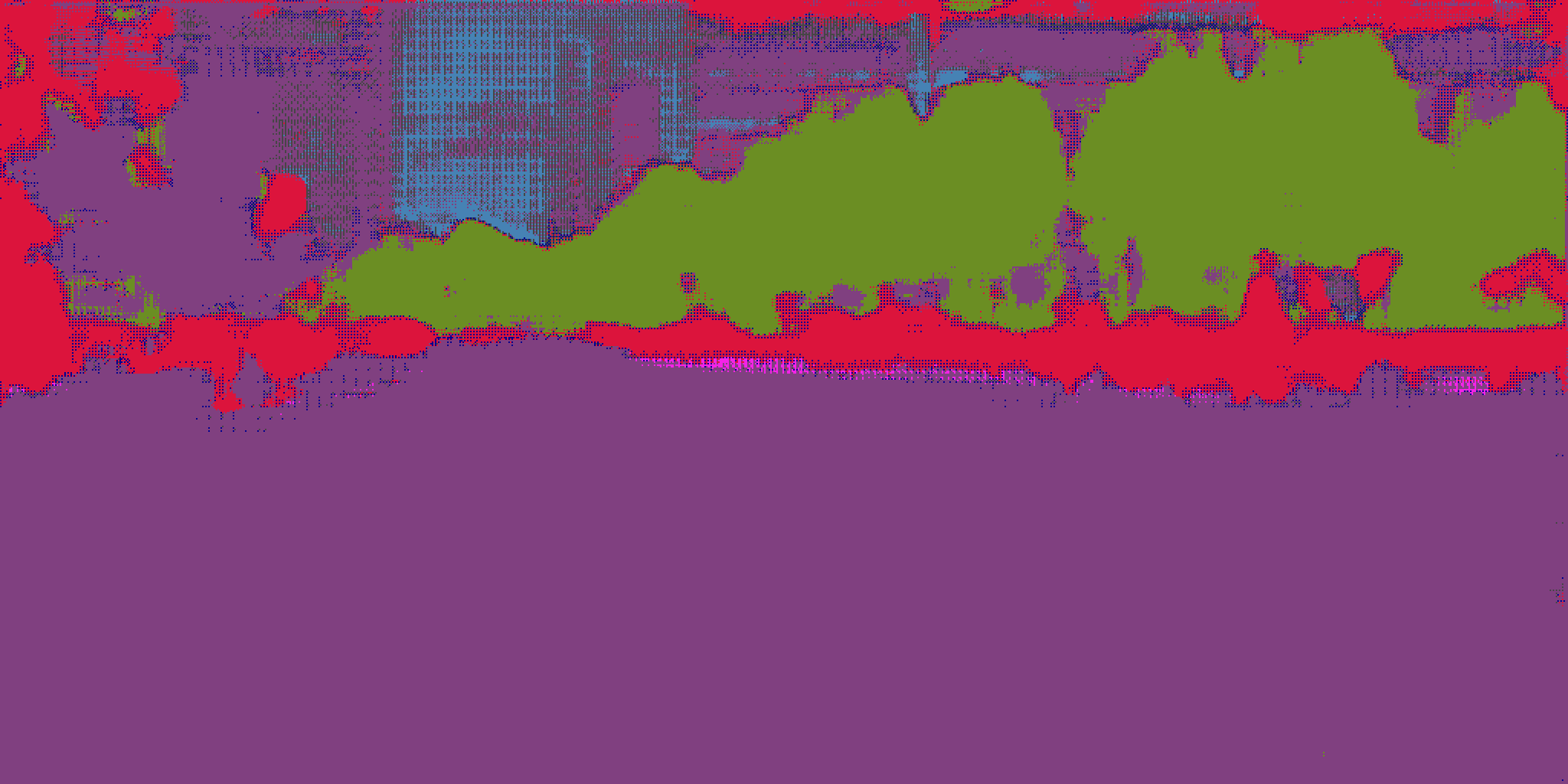}\vspace{0.1cm}
        \includegraphics[width=\linewidth]{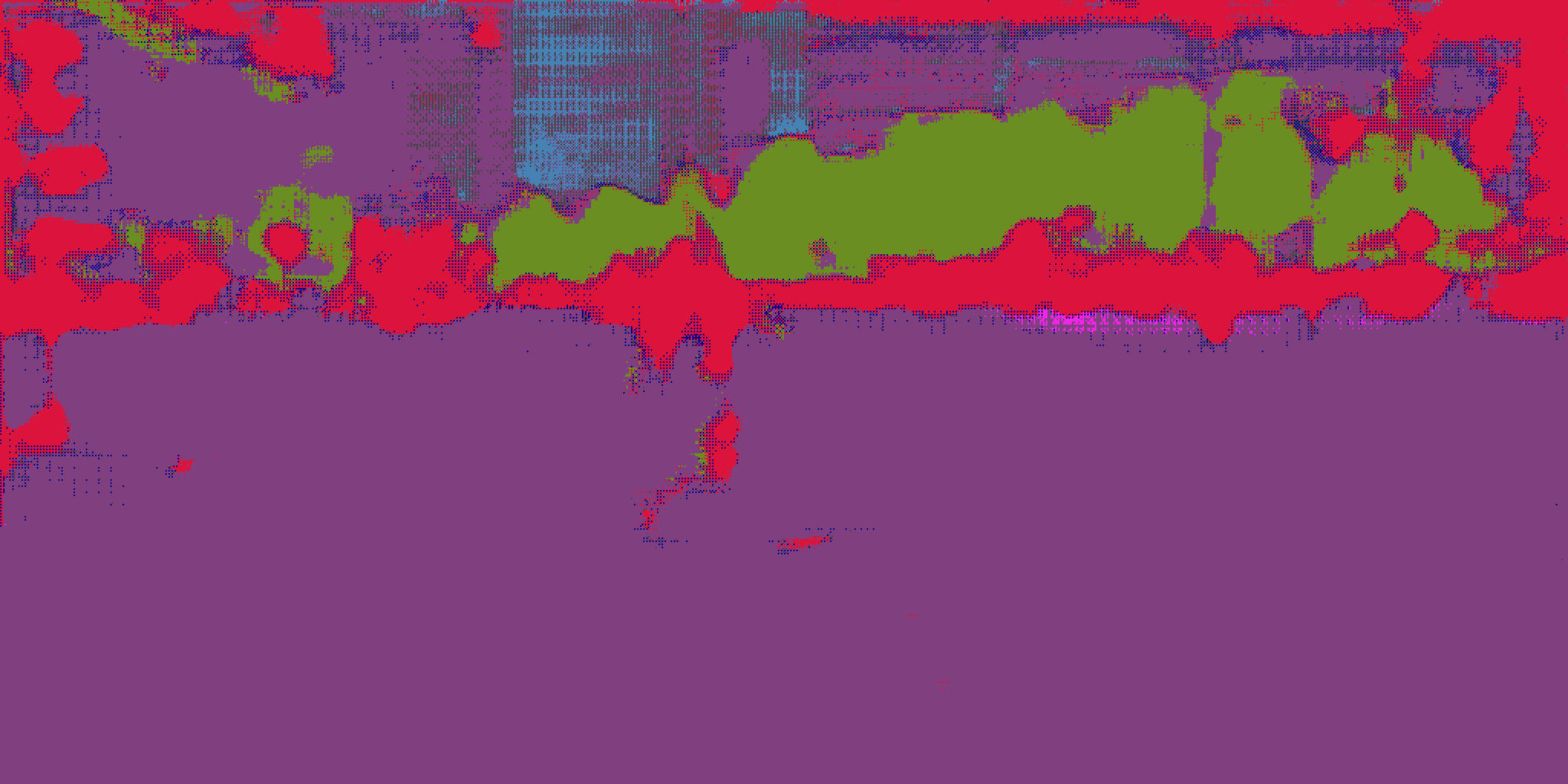}\vspace{0.1cm}
        \includegraphics[width=\linewidth]{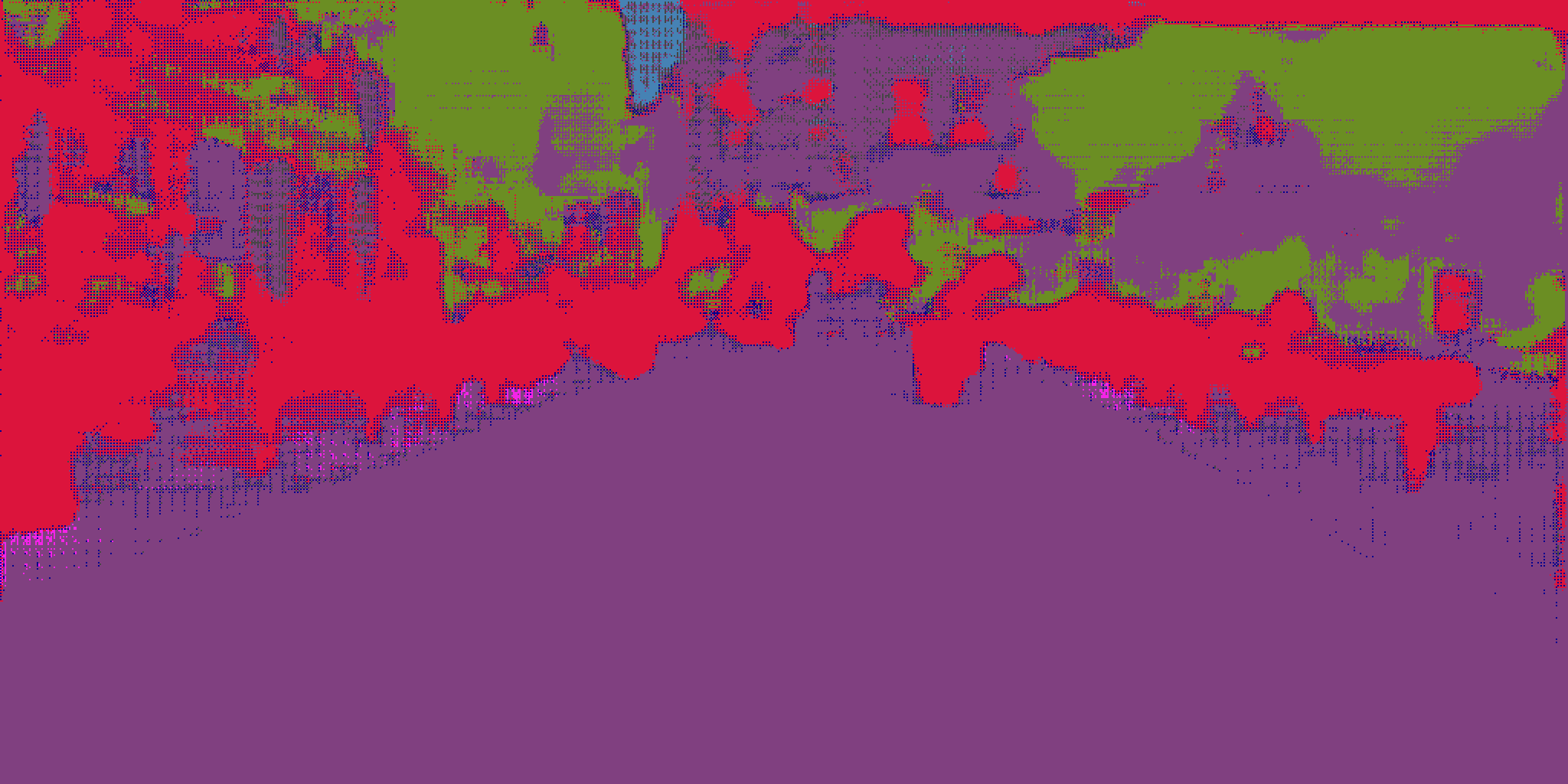}\vspace{0.1cm}
        \includegraphics[width=\linewidth]{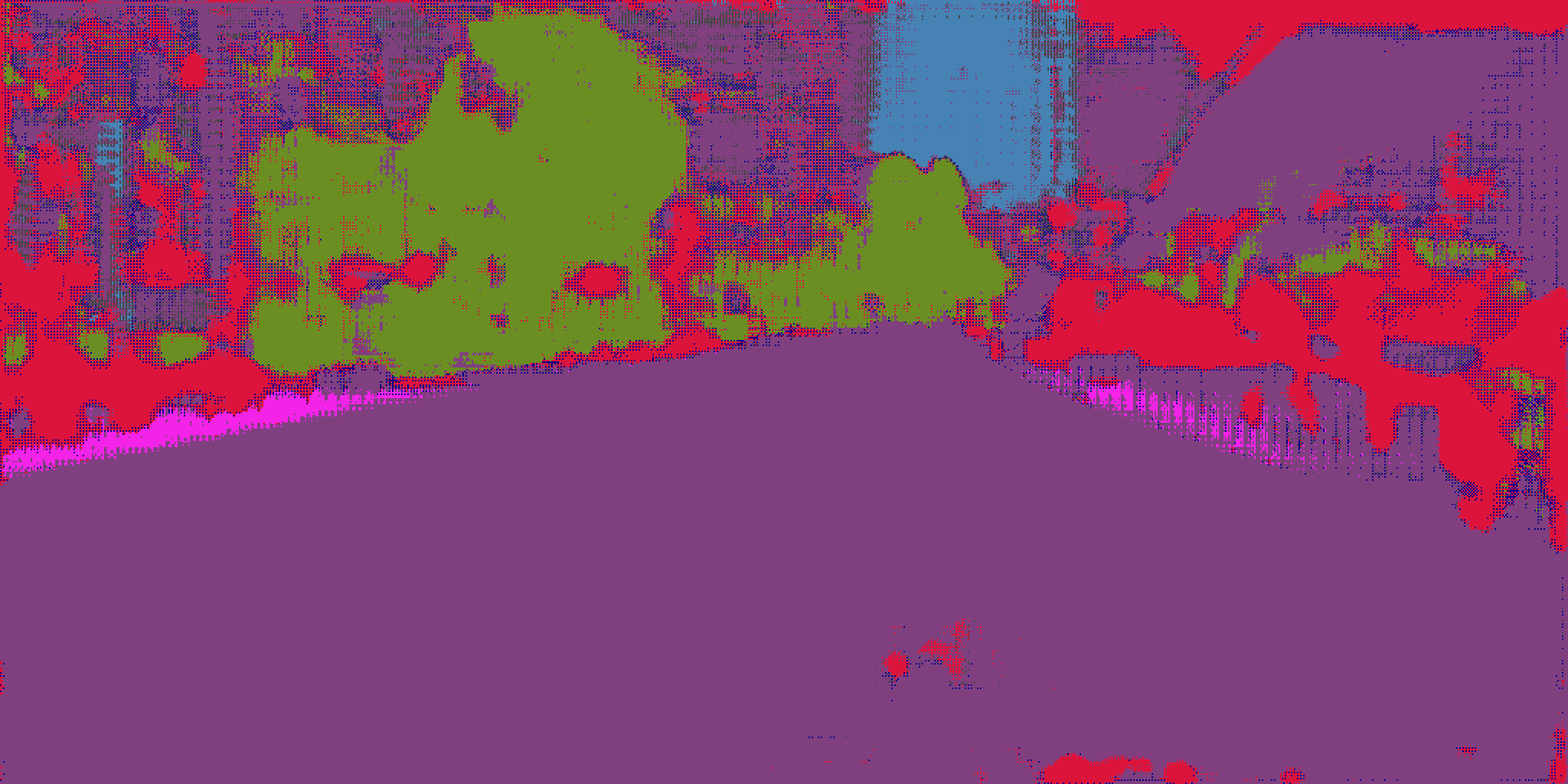}\vspace{0.1cm}
        \includegraphics[width=\linewidth]{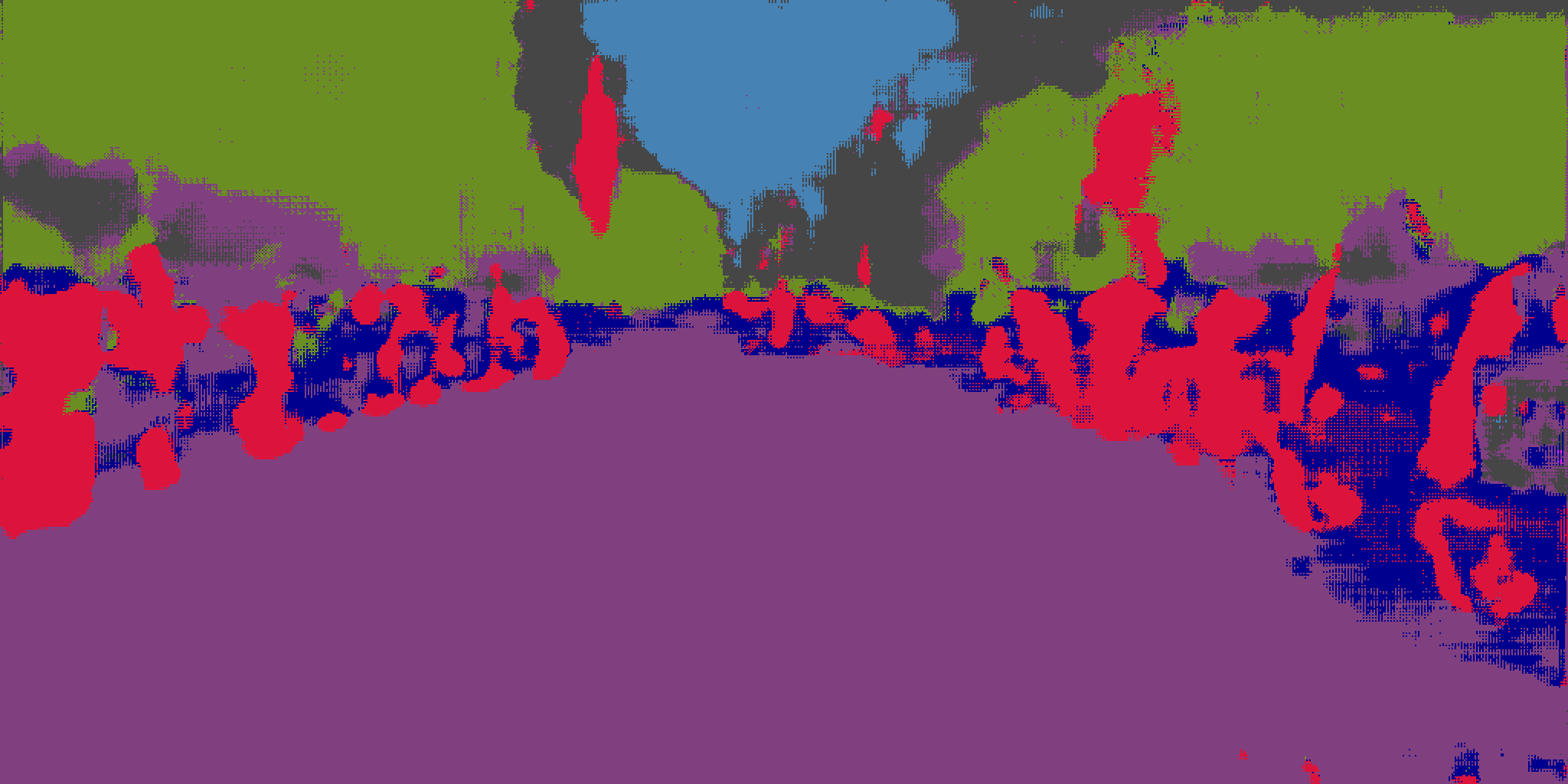}\vspace{0.1cm}
        \includegraphics[width=\linewidth]{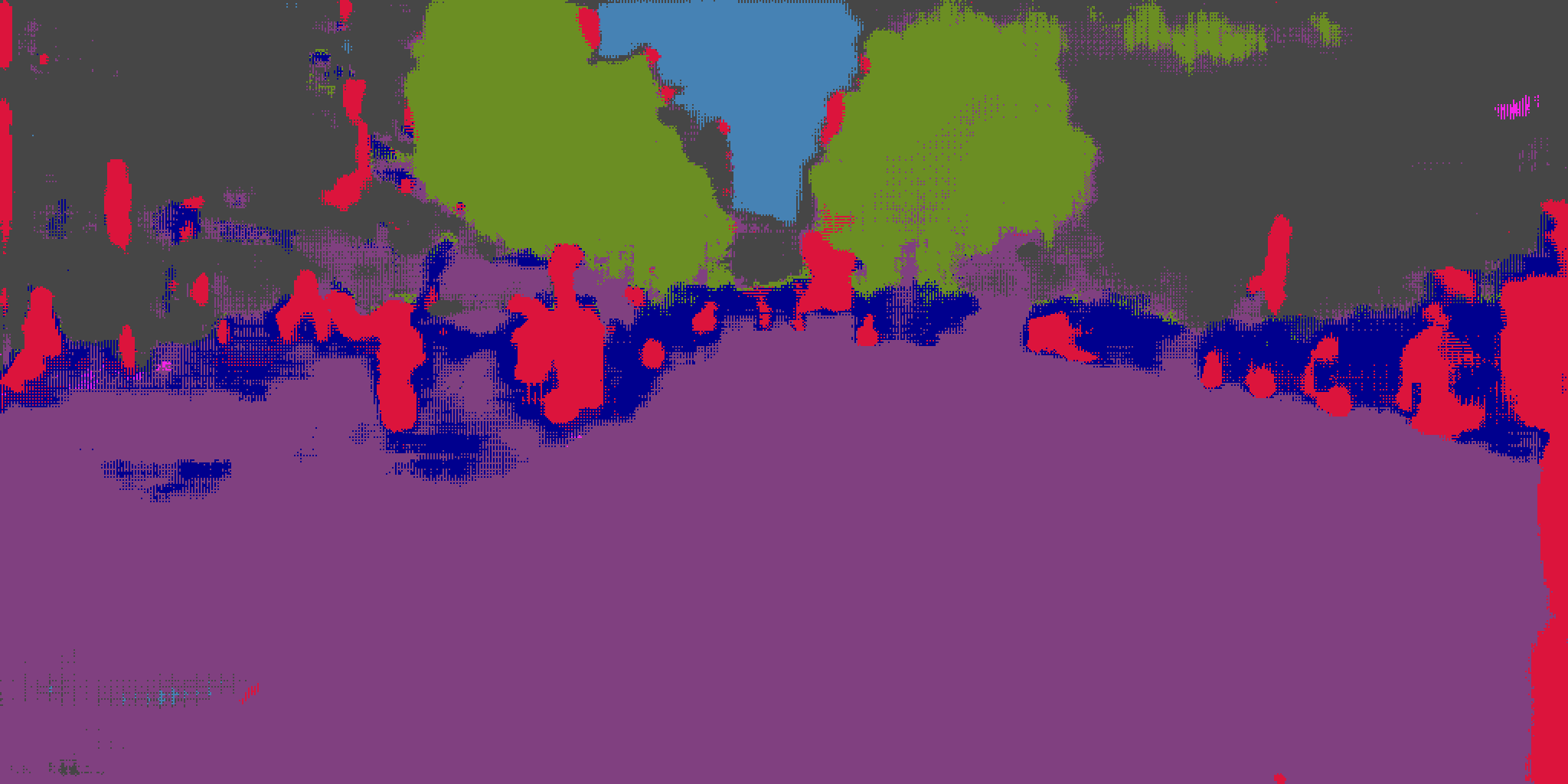}\vspace{0.1cm}
        \includegraphics[width=\linewidth]{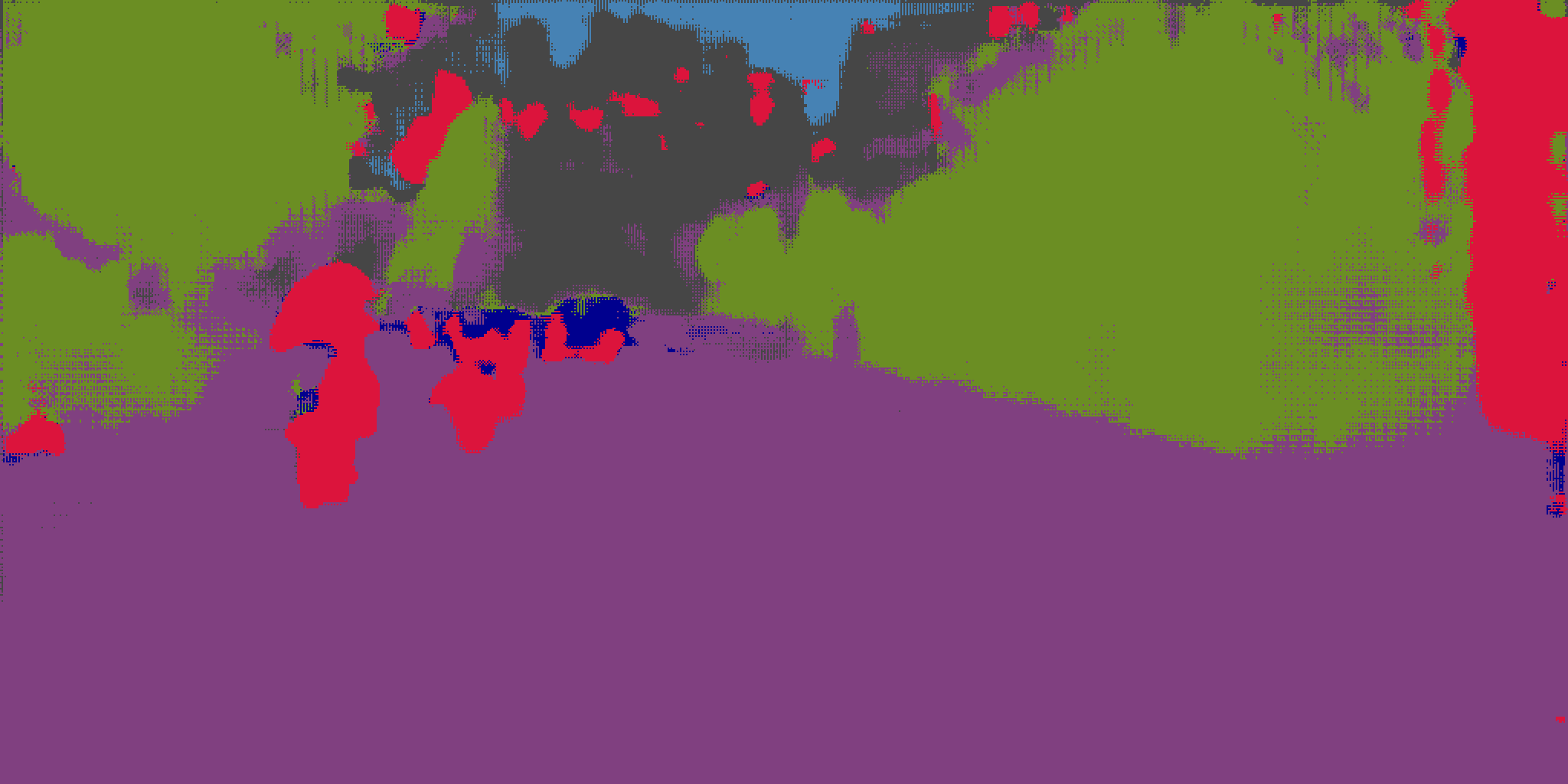}\vspace{0.1cm}
        \includegraphics[width=\linewidth]{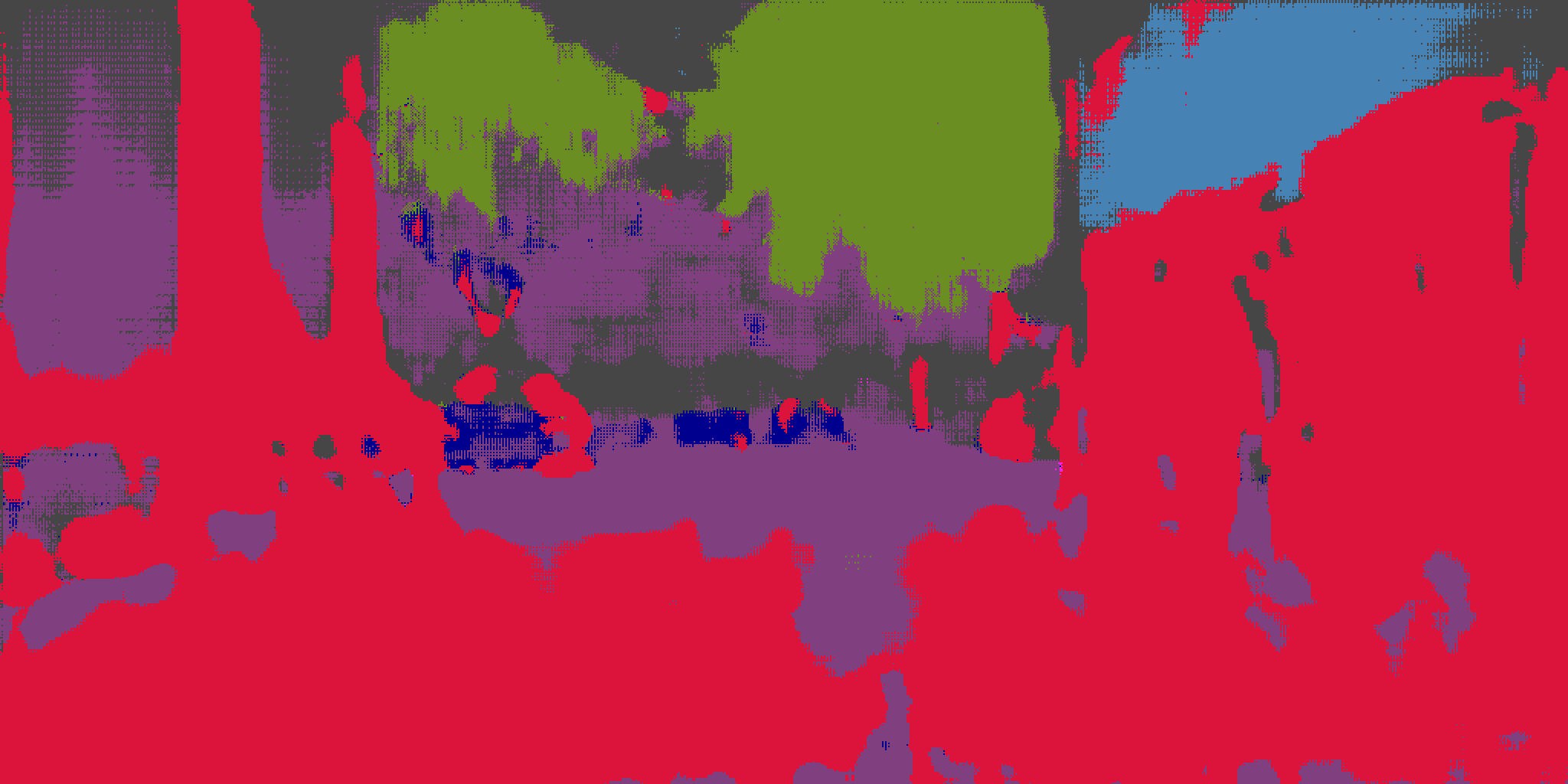}
    \end{minipage}\hfill
    \begin{minipage}{0.24\linewidth}
        \centering
        \caption*{Ours}
        \includegraphics[width=\linewidth]{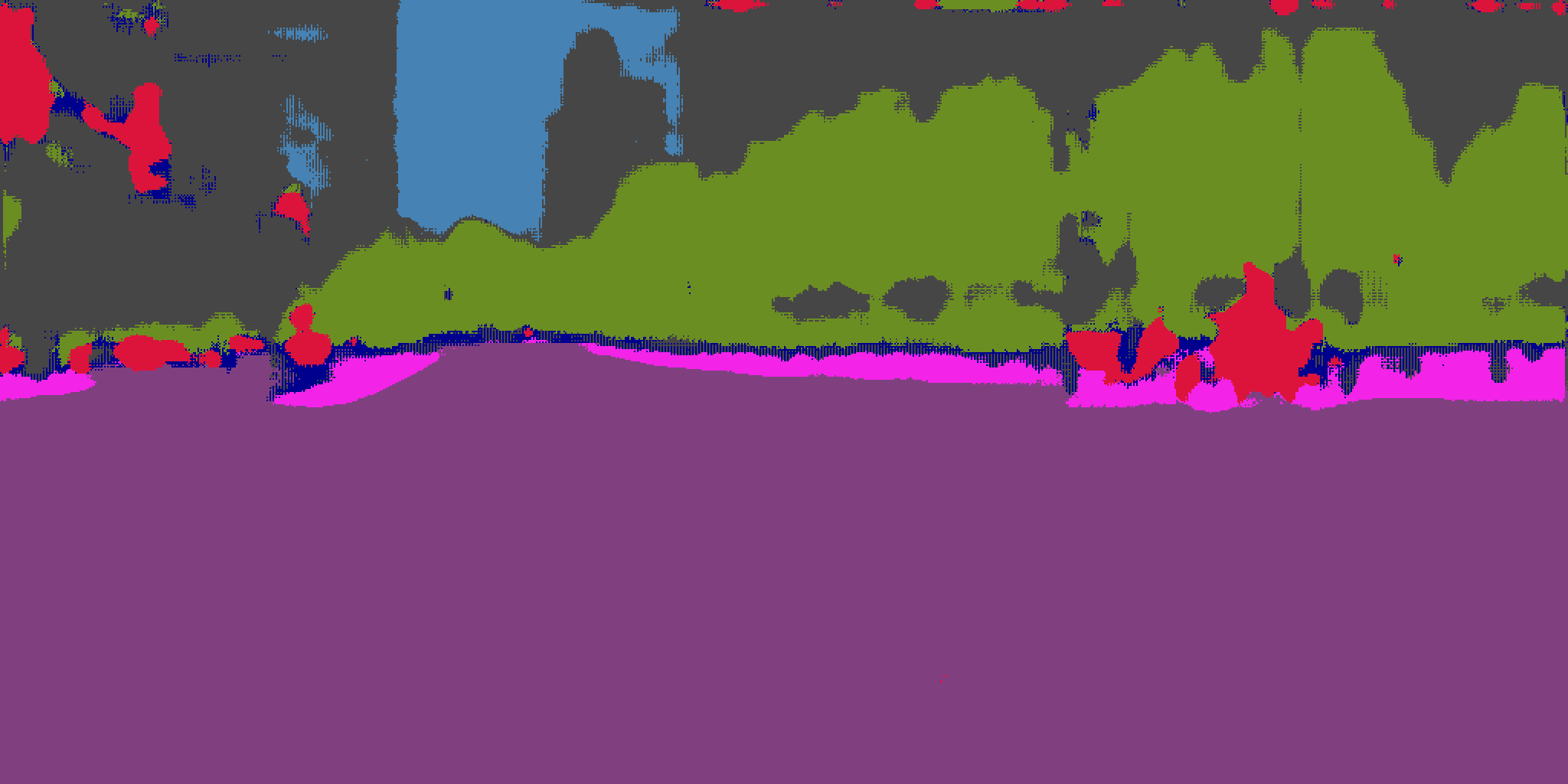}\vspace{0.1cm}
        \includegraphics[width=\linewidth]{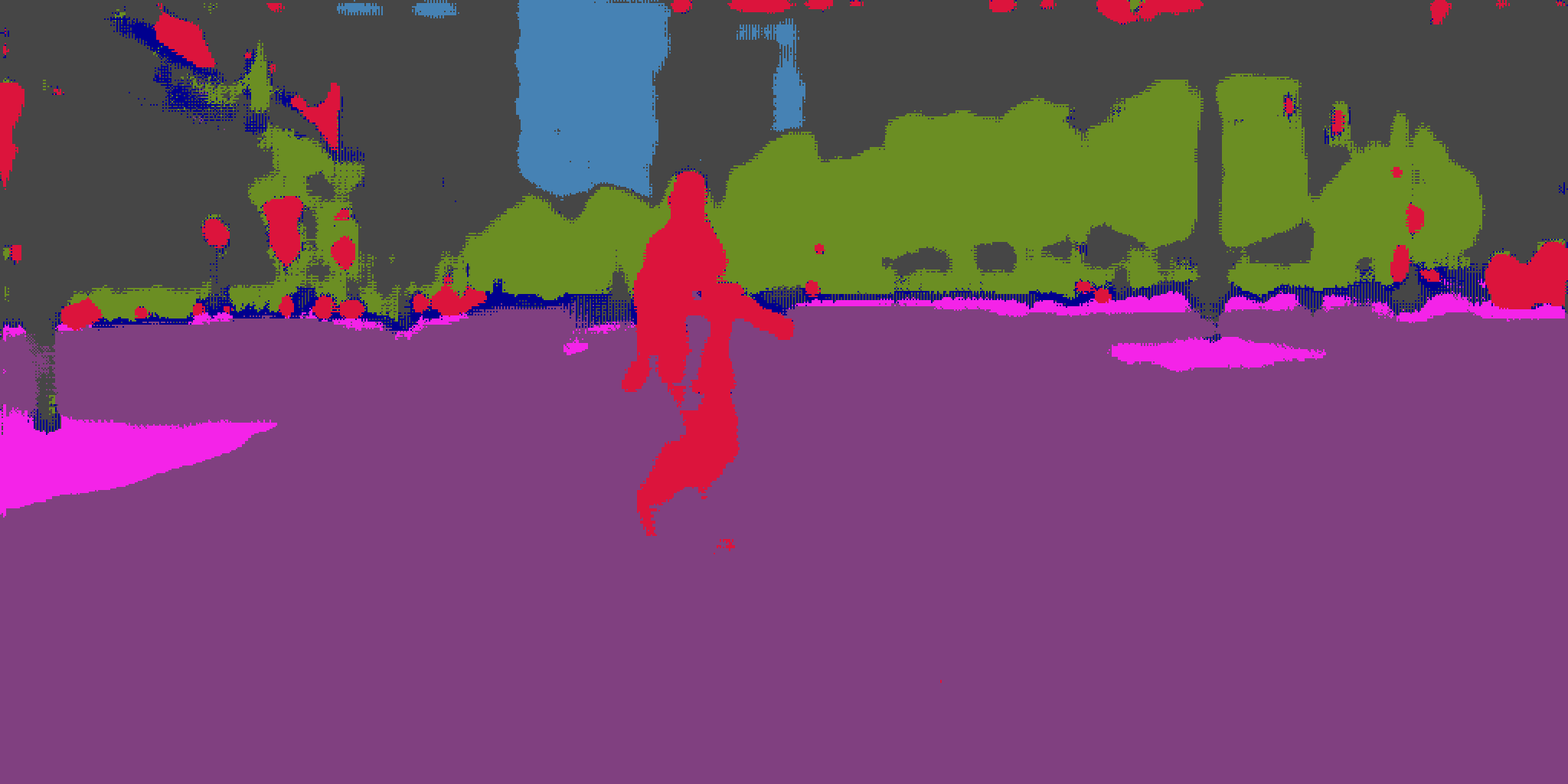}\vspace{0.1cm}
        \includegraphics[width=\linewidth]{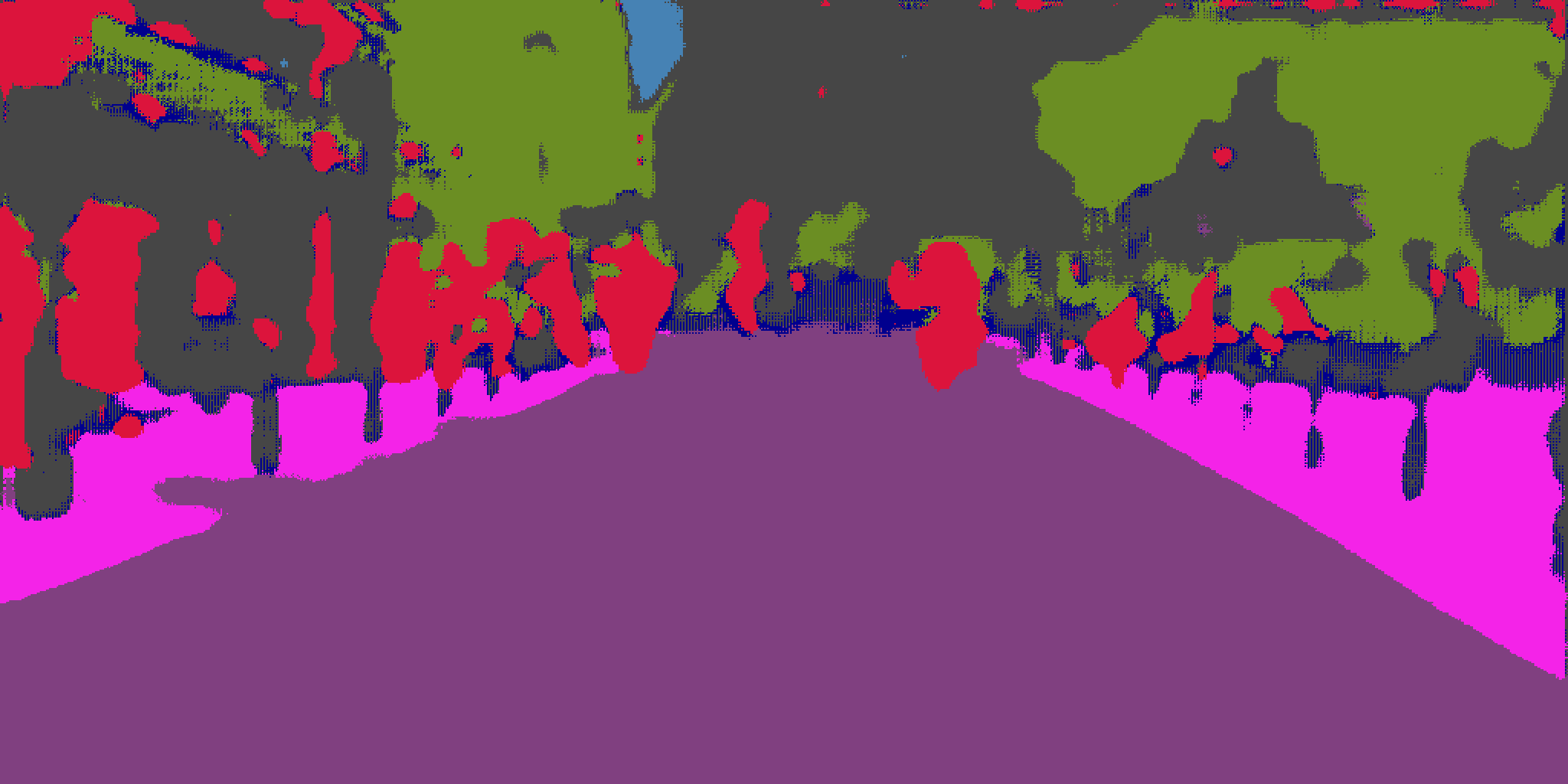}\vspace{0.1cm}
        \includegraphics[width=\linewidth]{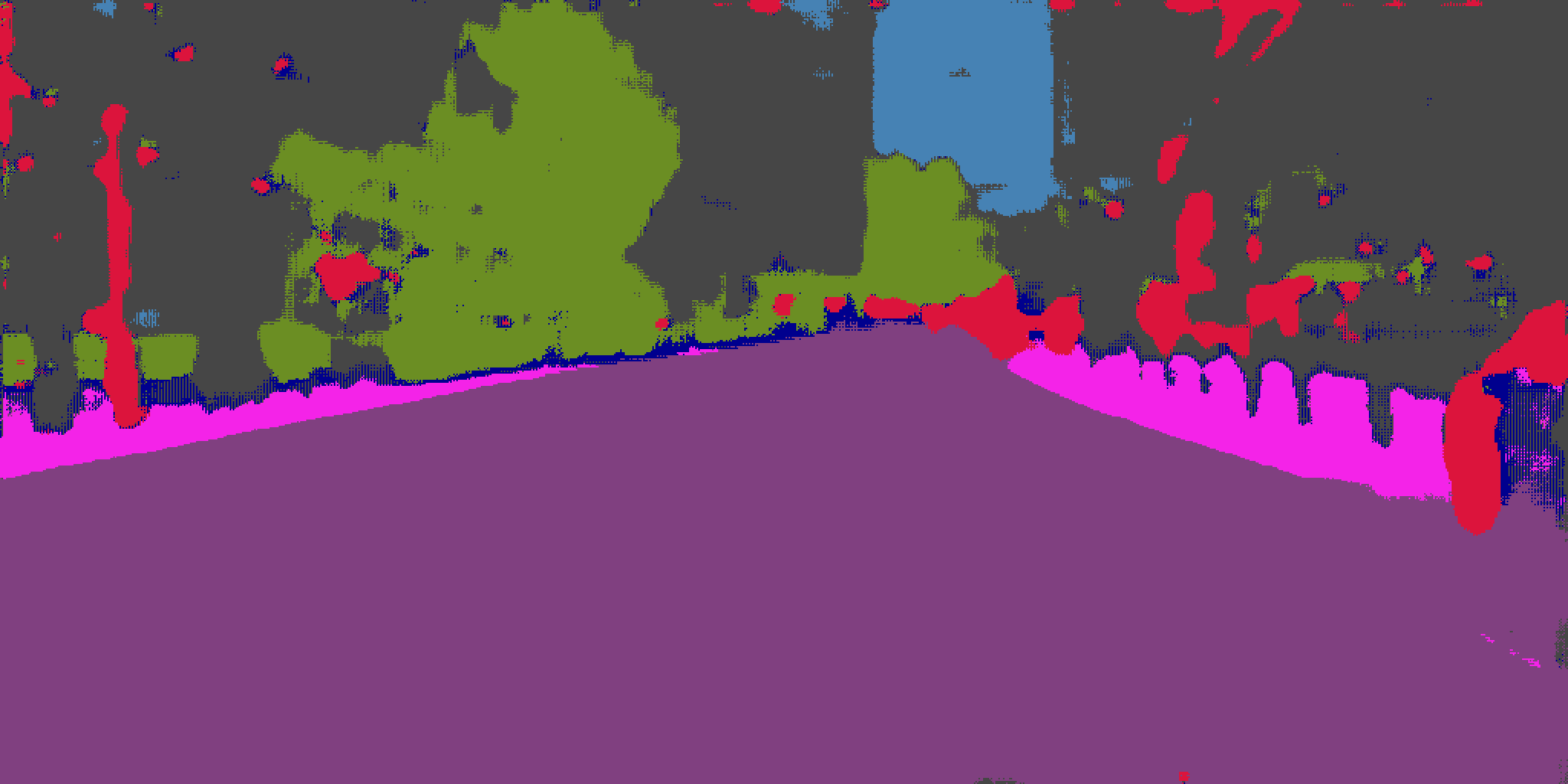}\vspace{0.1cm}
        \includegraphics[width=\linewidth]{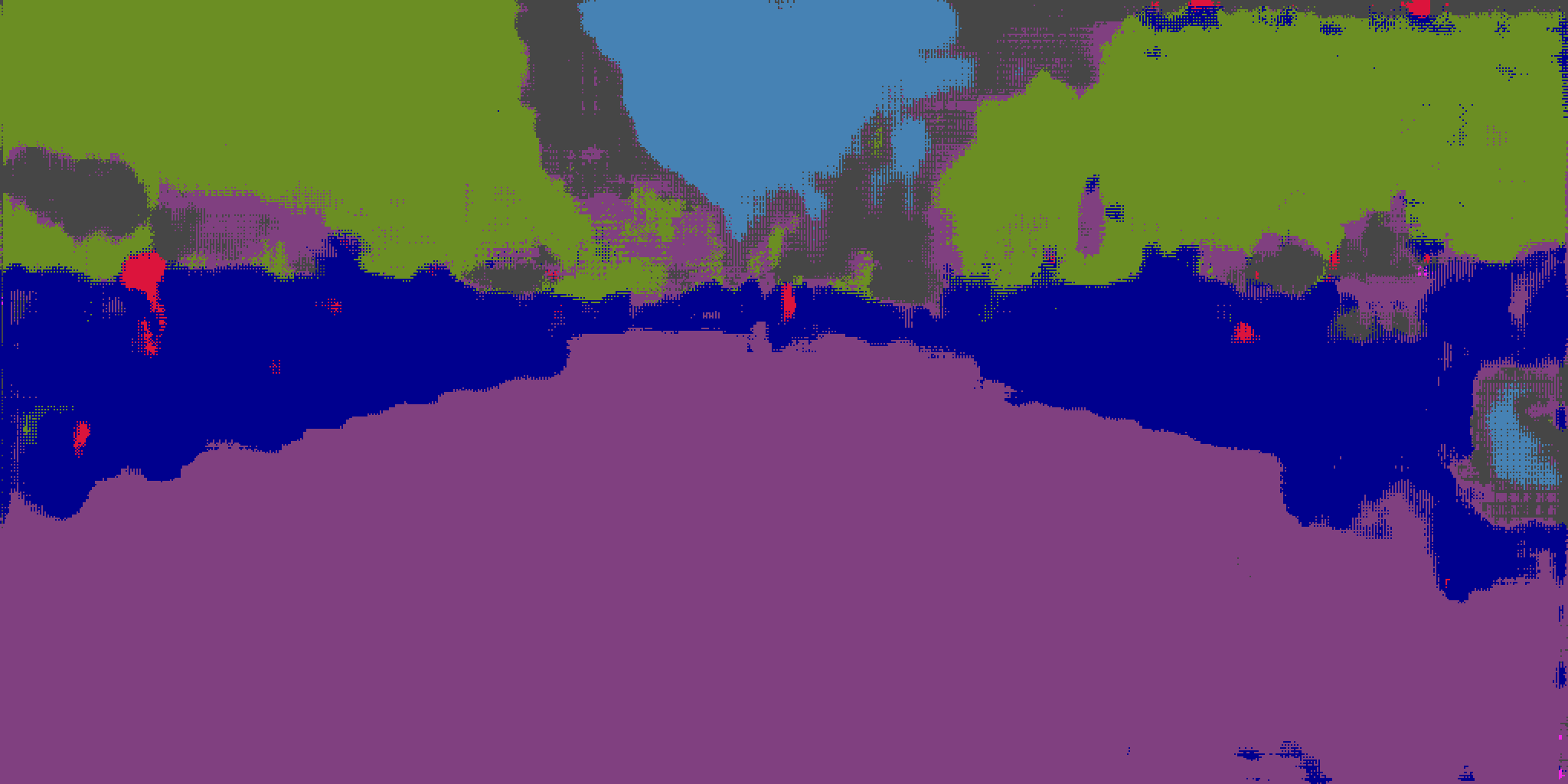}\vspace{0.1cm}
        \includegraphics[width=\linewidth]{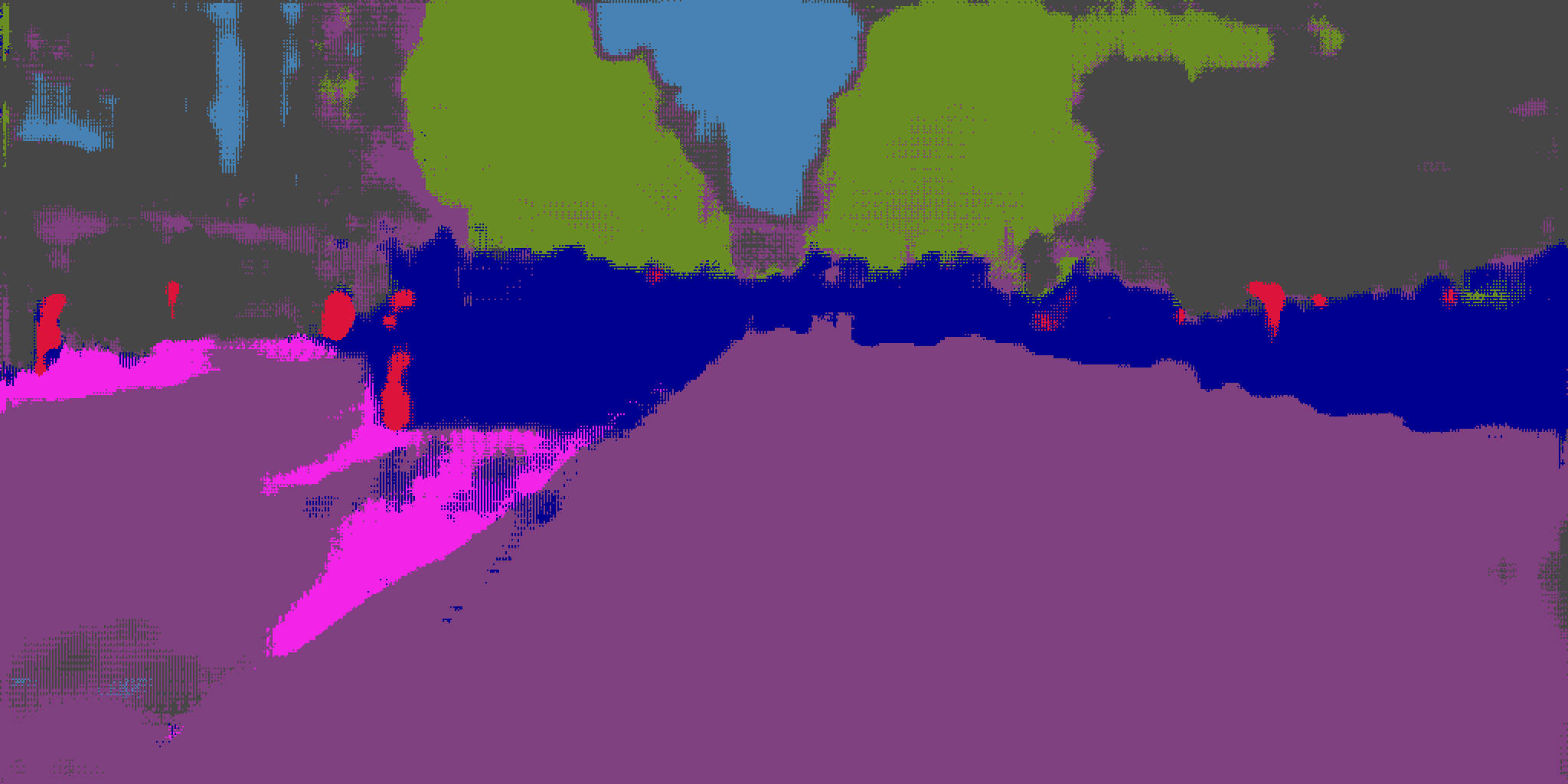}\vspace{0.1cm}
        \includegraphics[width=\linewidth]{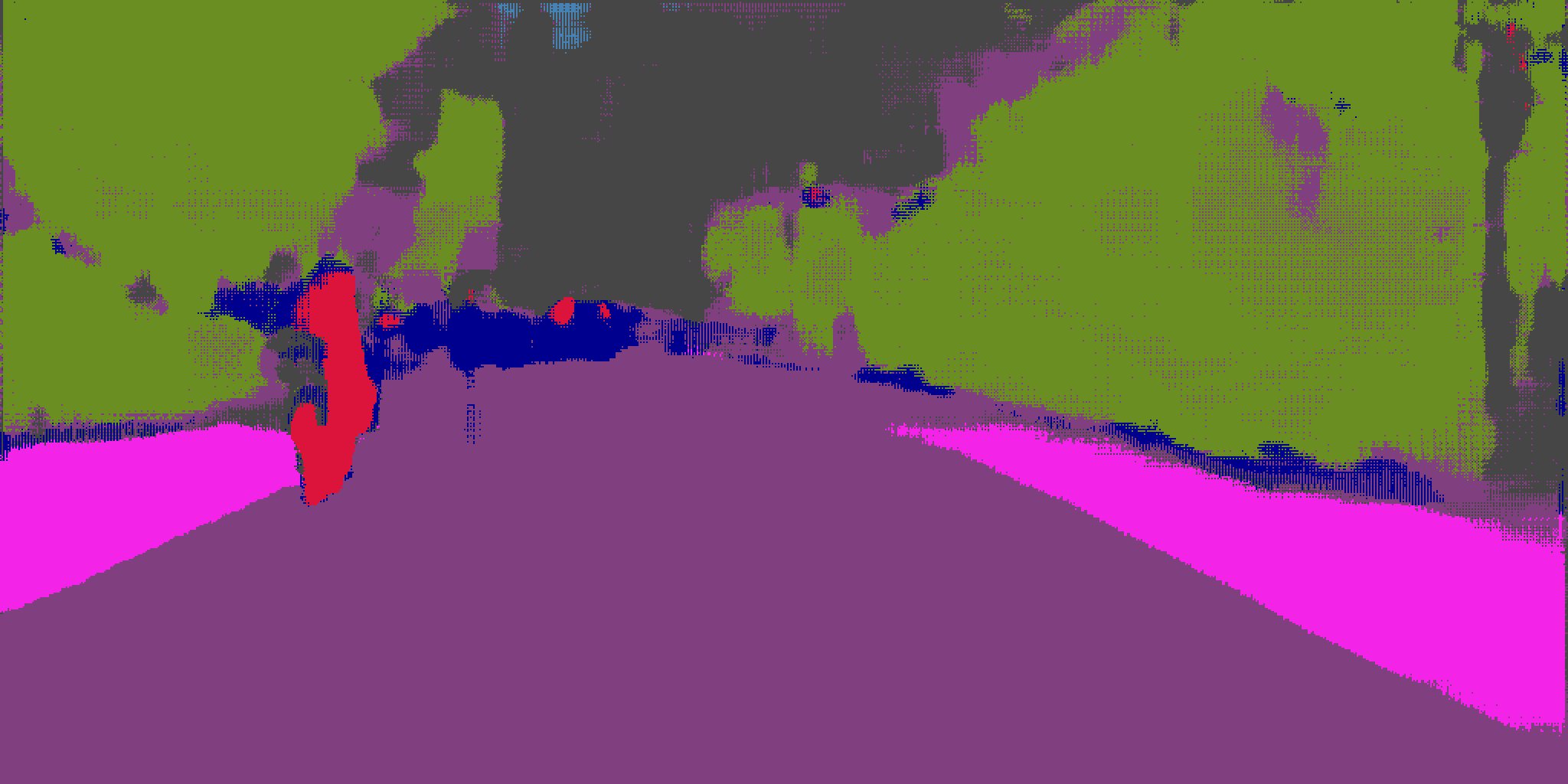}\vspace{0.1cm}
        \includegraphics[width=\linewidth]{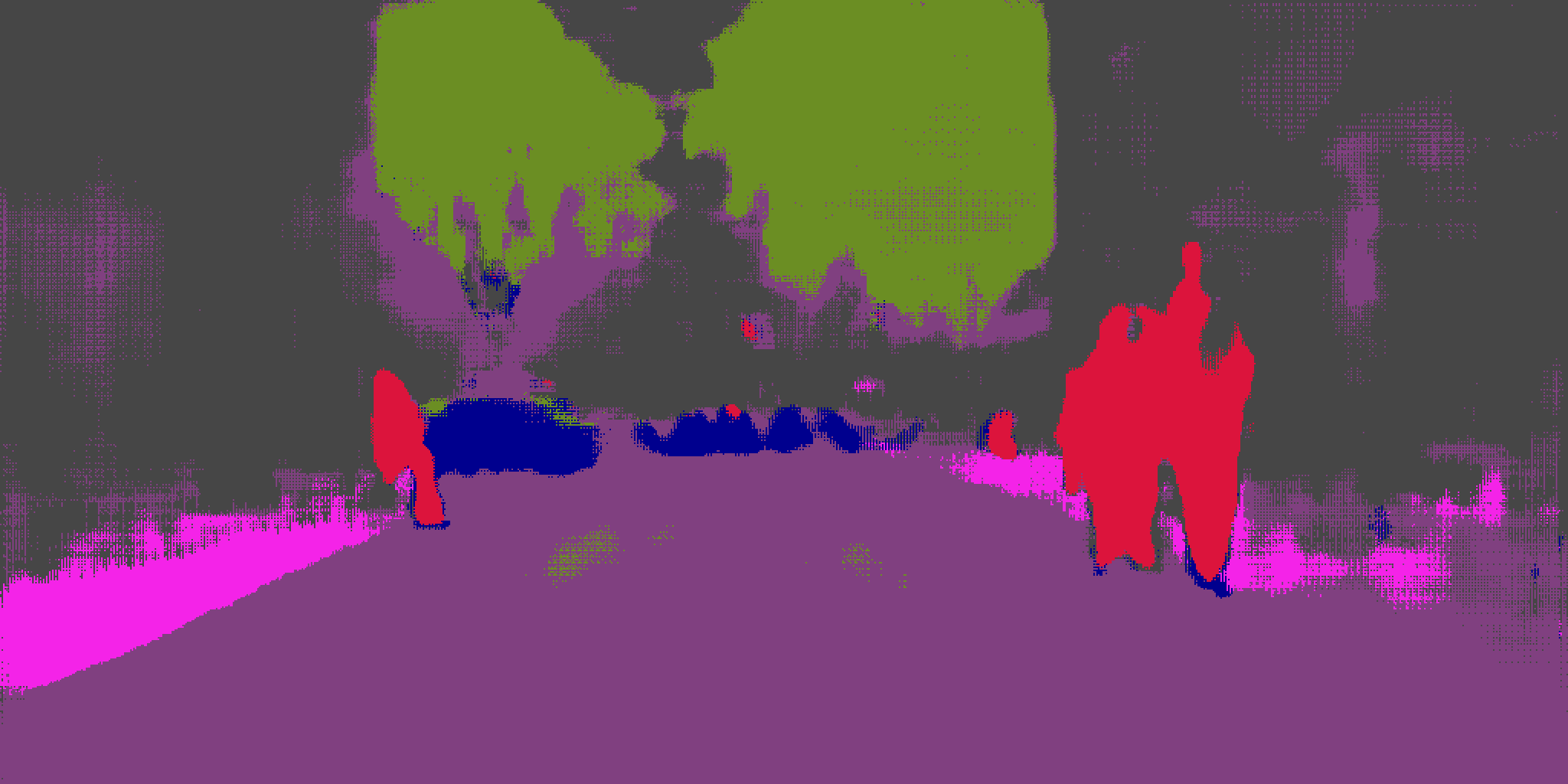}
    \end{minipage}
\caption{Example results on the \textbf{Cityscapes} validation set (the four rows from the top are results from 1-shot class-incremental learning, and the last four rows are results from 5-shot learning). Left: Input image; Middle left: Ground-truth; Middle right: Prediction from the baseline method (FT+KD); Right: Prediction from our method (FT+KD+PL).}
\label{img:city}
\end{figure*}

\subsection{Experimental Results}
\label{sec:exp:res}

The quantitative results we obtained on the Cityscapes and KITTI datasets are presented in Table~\ref{tab:cityscapes} and Table~\ref{tab:kitti} respectively.
We name the methods by the components being used. Specifically, FT, KD and PL refer to Fine-Tuning, Knowledge Distillation and Pseudo-Labeling, respectively. We present two competing methods: one with learning the first and the third terms in Equation~\ref{eqn:learn} (FT+KD) and one with learning all the terms in Equation~\ref{eqn:learn} (FT+KD+PL).
We note that learning without knowledge distillation produces very poor results, as the method will inevitably suffer from catastrophic forgetting.

\noindent \textbf{Results on Cityscapes.} We can clearly see from Table \ref{tab:cityscapes} that the pseudo-labeling strategy we proposed can effectively make use of the unlabeled data and boost performance. In the vast majority of training stages, incorporating PL leads to significant performance improvements. Specifically, in the 5-shot class-incremental setting, adding PL results in an average performance gain of 2.1\%, 3.3\%, 2.3\%, 6.1\%, and 3.3\% for Task 1, Task 2, Task $1 \cup 2$, Task 3 and Task $1 \cup 2 \cup 3$, respectively. We present some qualitative results in Figure~\ref{img:city}.

\begin{figure*}[t!]
    \centering
    \begin{minipage}{0.24\linewidth}
        \centering
        \captionsetup{labelformat=empty}
        \caption*{Image}
        \includegraphics[width=\linewidth]{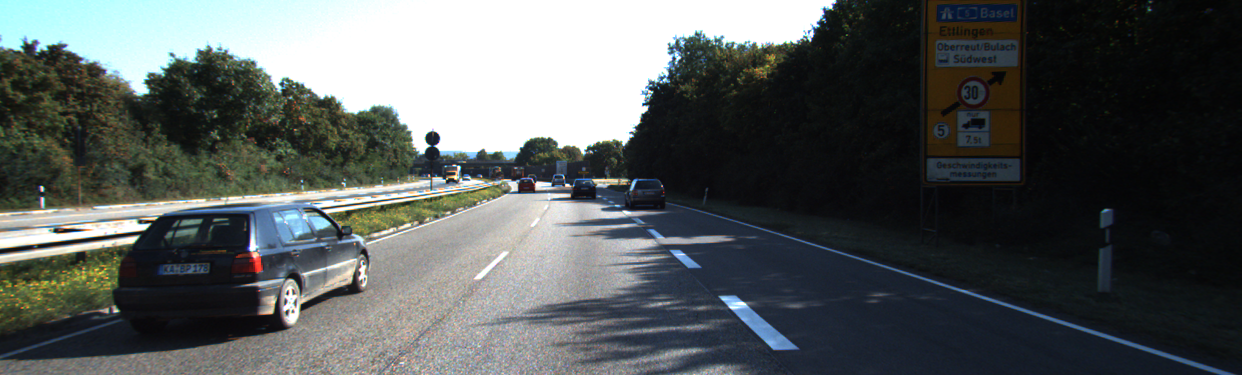}\vspace{0.1cm}
        \includegraphics[width=\linewidth]{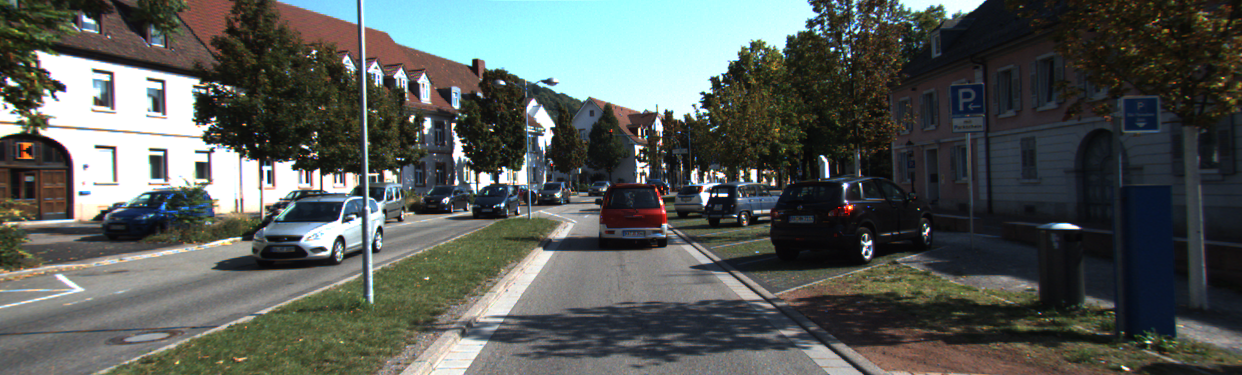}\vspace{0.1cm}
        \includegraphics[width=\linewidth]{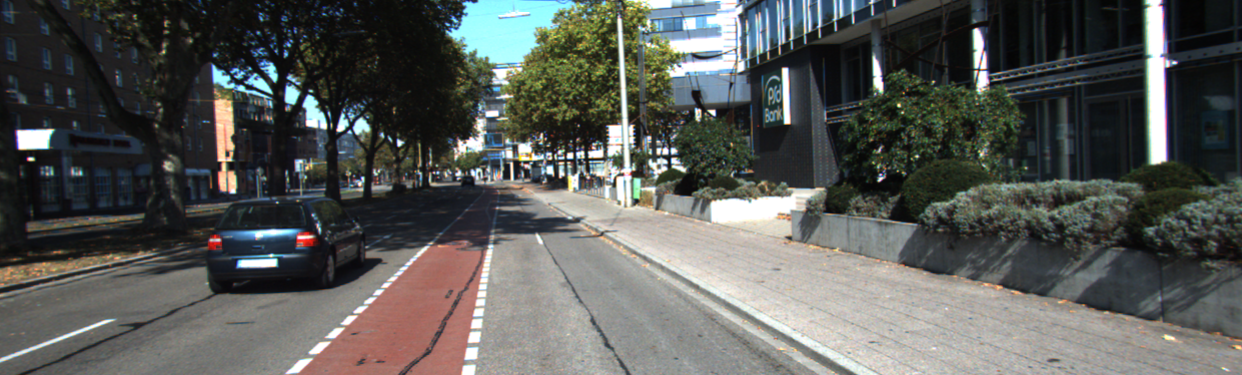}\vspace{0.1cm}
        \includegraphics[width=\linewidth]{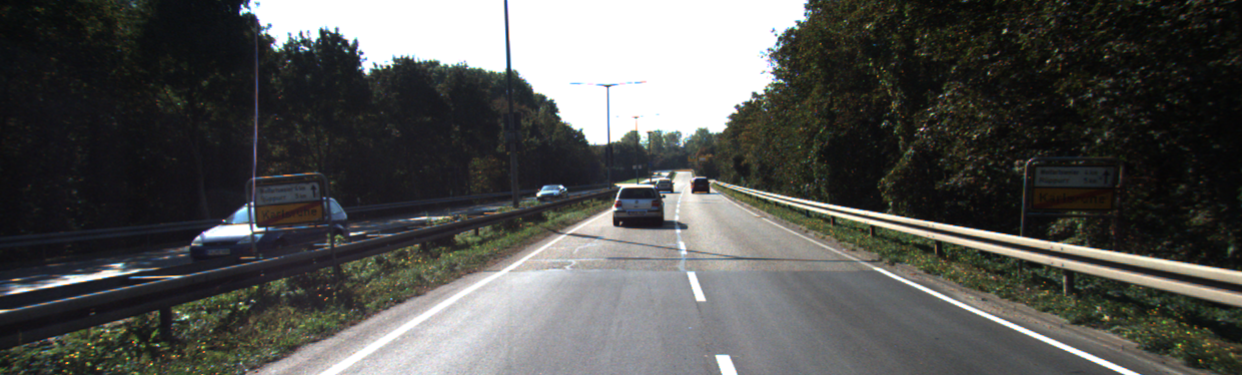}\vspace{0.1cm}
        \includegraphics[width=\linewidth]{kitti/images/kitti_000100_10_leftImg8bit.png}\vspace{0.1cm}
        \includegraphics[width=\linewidth]{kitti/images/kitti_000123_10_leftImg8bit.png}\vspace{0.1cm}
        \includegraphics[width=\linewidth]{kitti/images/kitti_000172_10_leftImg8bit.png}\vspace{0.1cm}
        \includegraphics[width=\linewidth]{kitti/images/kitti_000179_10_leftImg8bit.png}
    \end{minipage}\hfill
    \begin{minipage}{0.24\linewidth}
        \centering
        \caption*{GT}
        \includegraphics[width=\linewidth]{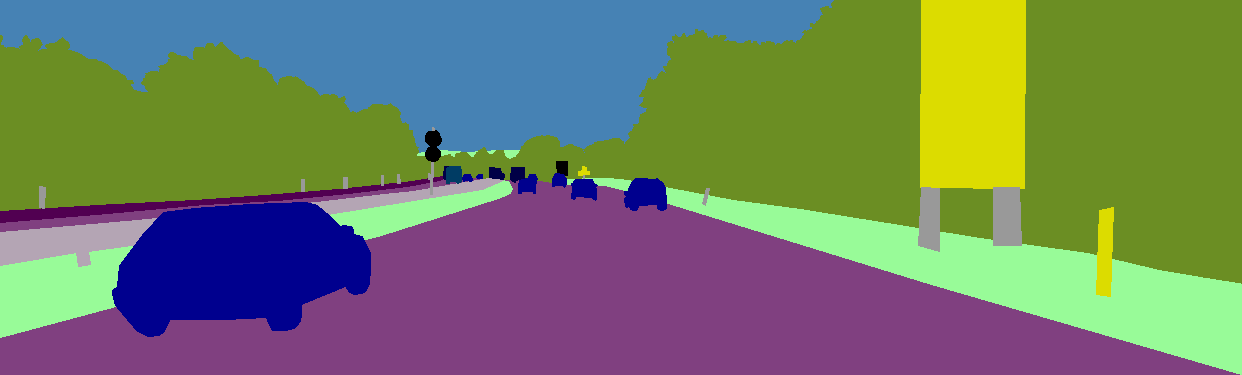}\vspace{0.1cm}
        \includegraphics[width=\linewidth]{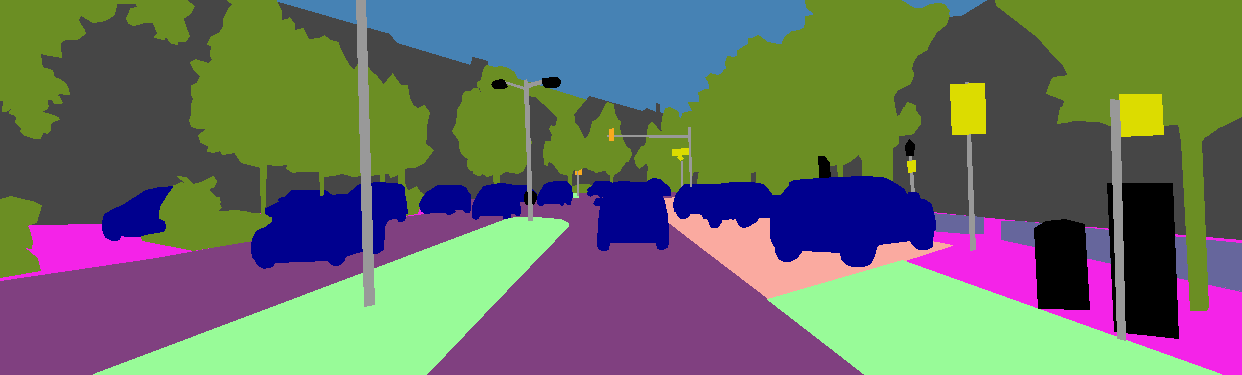}\vspace{0.1cm}
        \includegraphics[width=\linewidth]{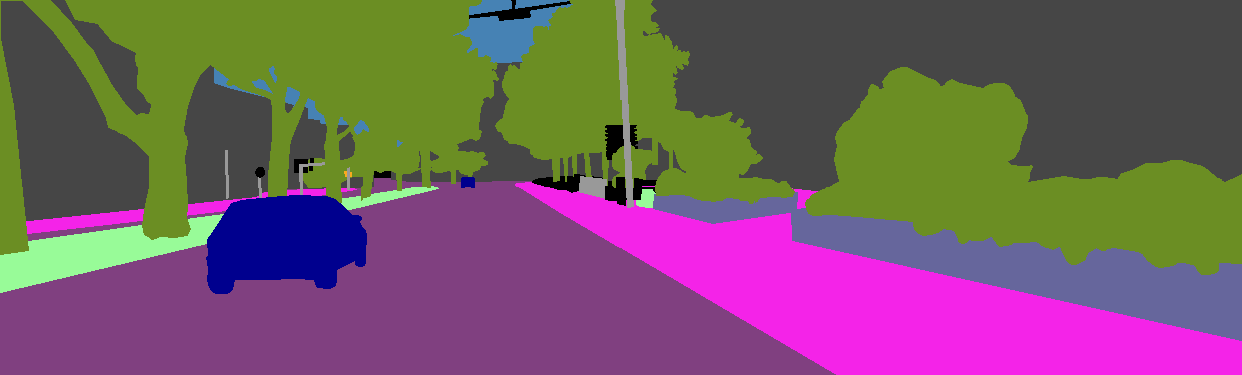}\vspace{0.1cm}
        \includegraphics[width=\linewidth]{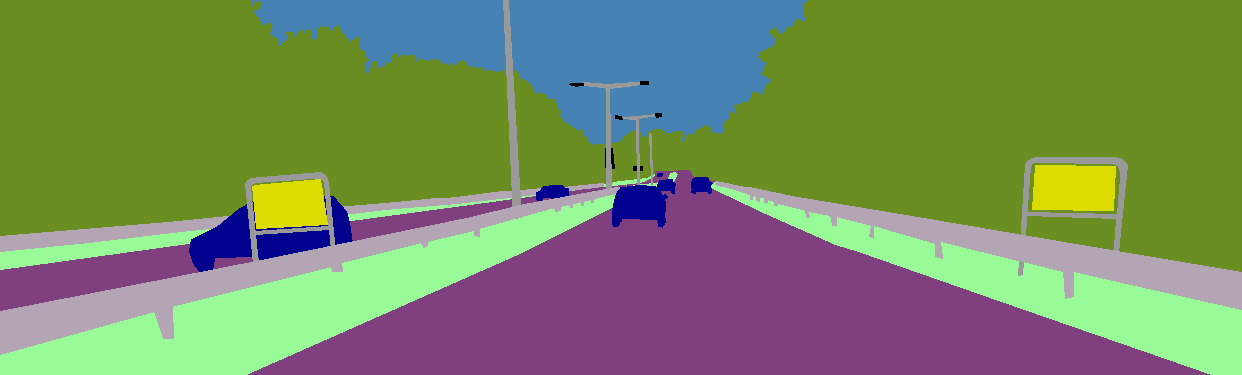}\vspace{0.1cm}
        \includegraphics[width=\linewidth]{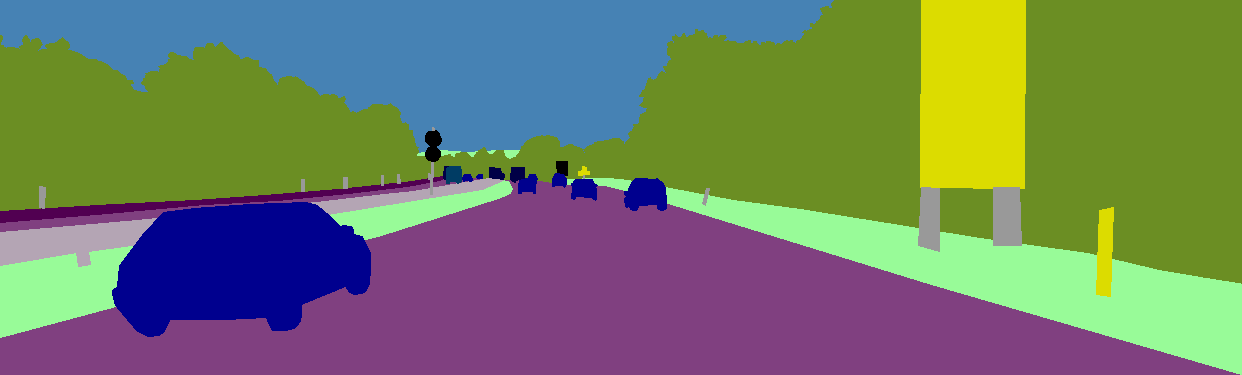}\vspace{0.1cm}
        \includegraphics[width=\linewidth]{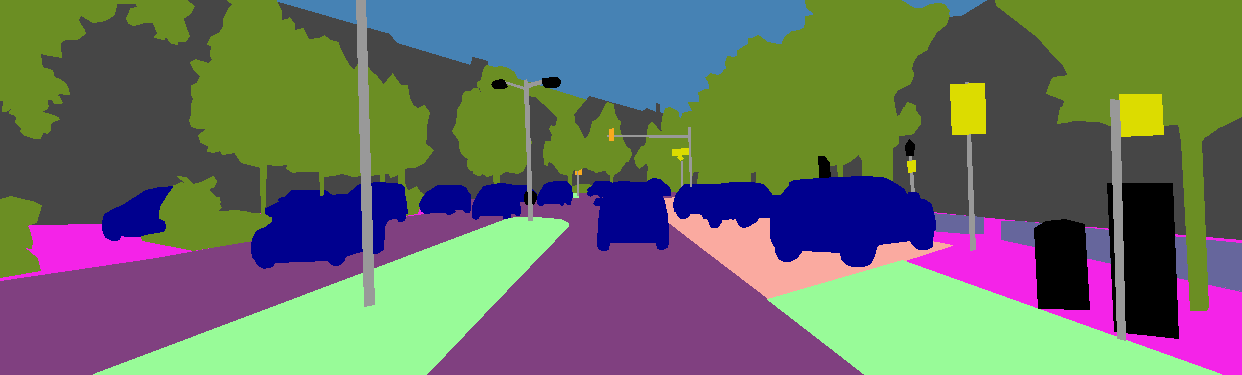}\vspace{0.1cm}
        \includegraphics[width=\linewidth]{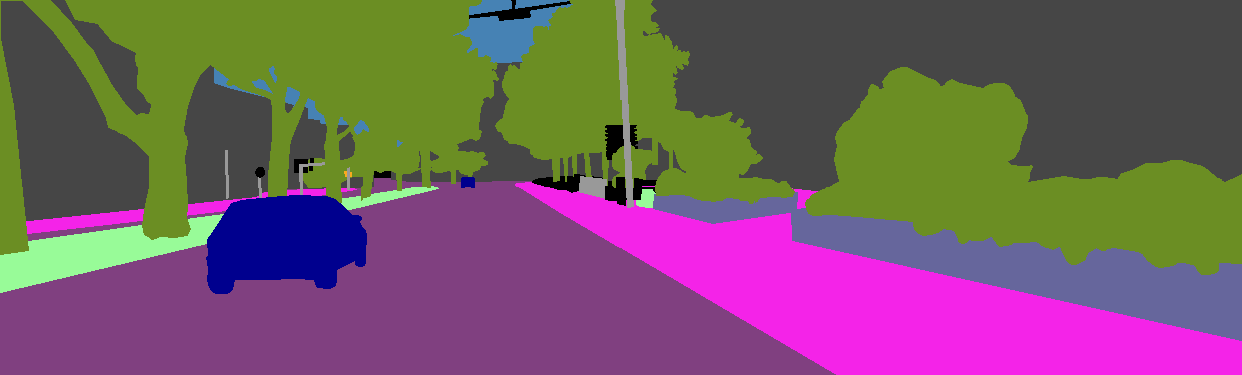}\vspace{0.1cm}
        \includegraphics[width=\linewidth]{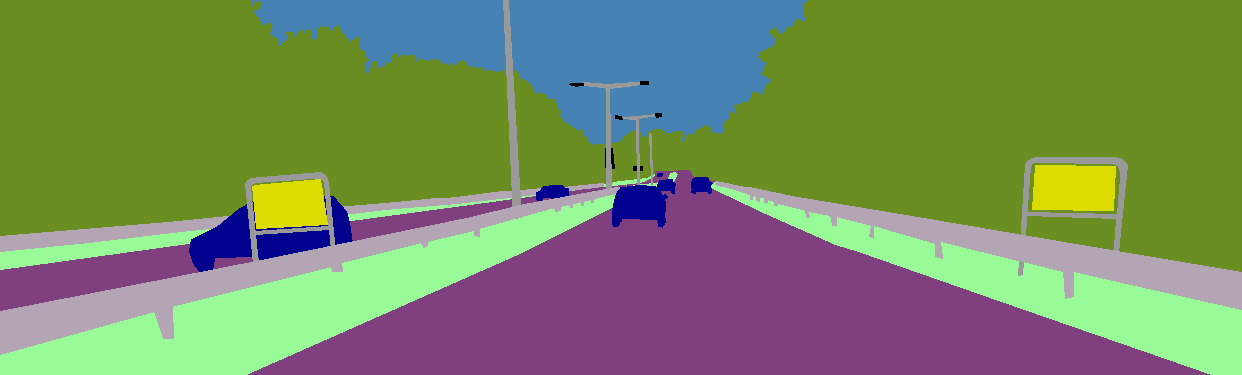}
    \end{minipage}\hfill
    \begin{minipage}{0.24\linewidth}
        \centering
        \caption*{Baseline}
        \includegraphics[width=\linewidth]{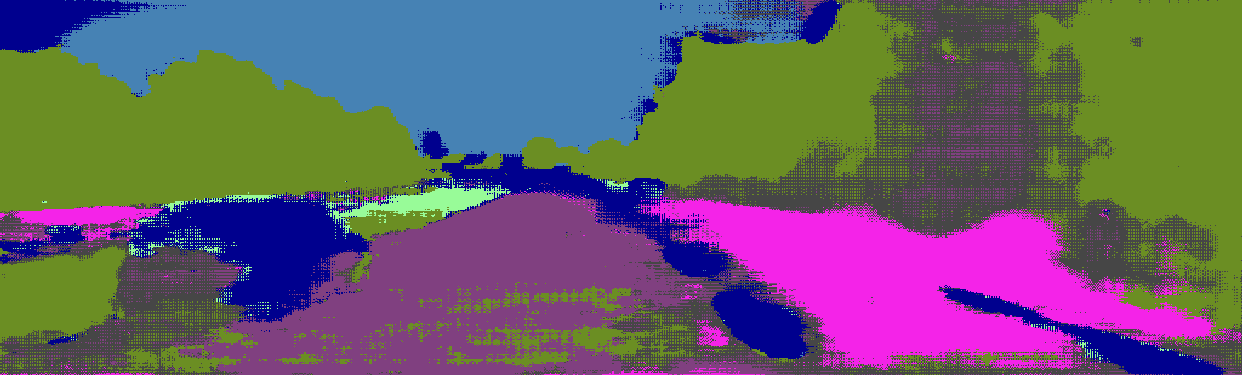}\vspace{0.1cm}
        \includegraphics[width=\linewidth]{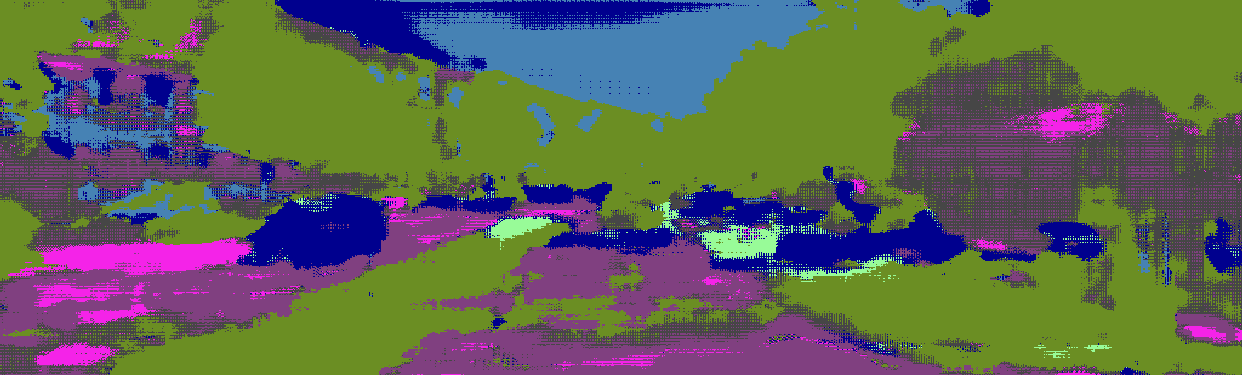}\vspace{0.1cm}
        \includegraphics[width=\linewidth]{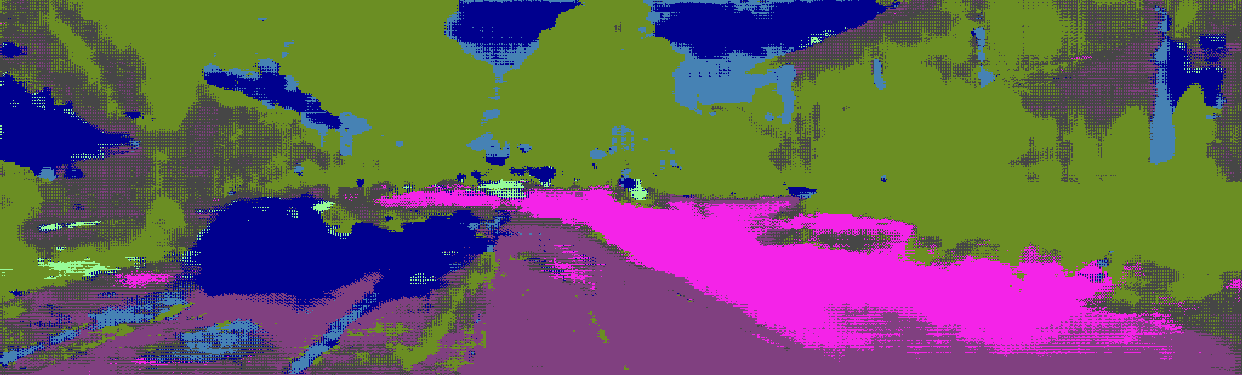}\vspace{0.1cm}
        \includegraphics[width=\linewidth]{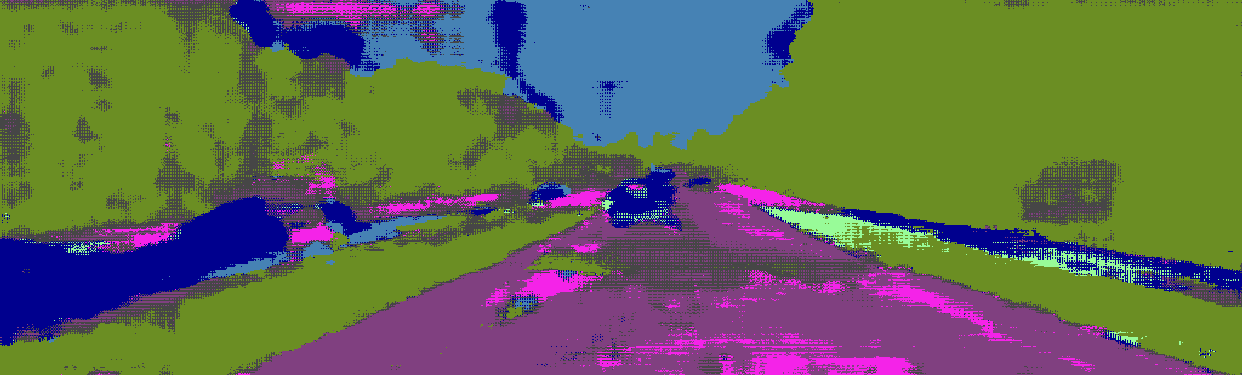}\vspace{0.1cm}
        \includegraphics[width=\linewidth]{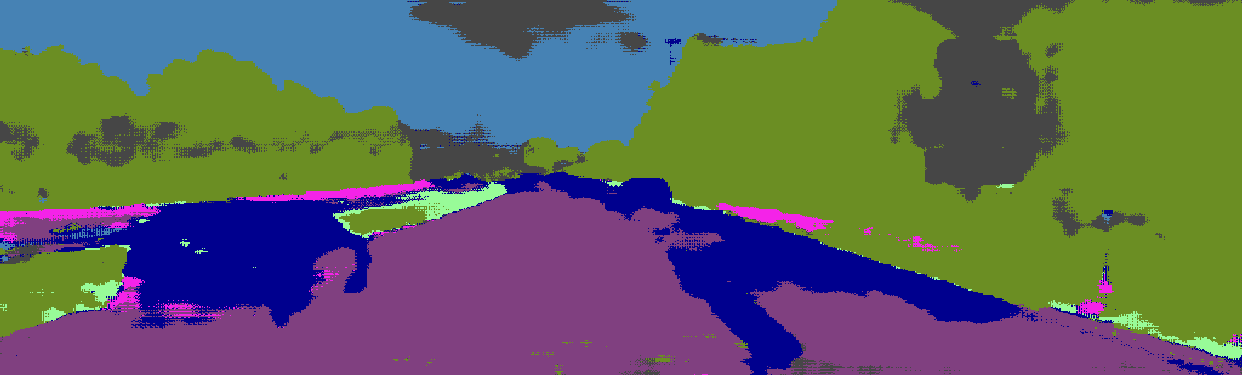}\vspace{0.1cm}
        \includegraphics[width=\linewidth]{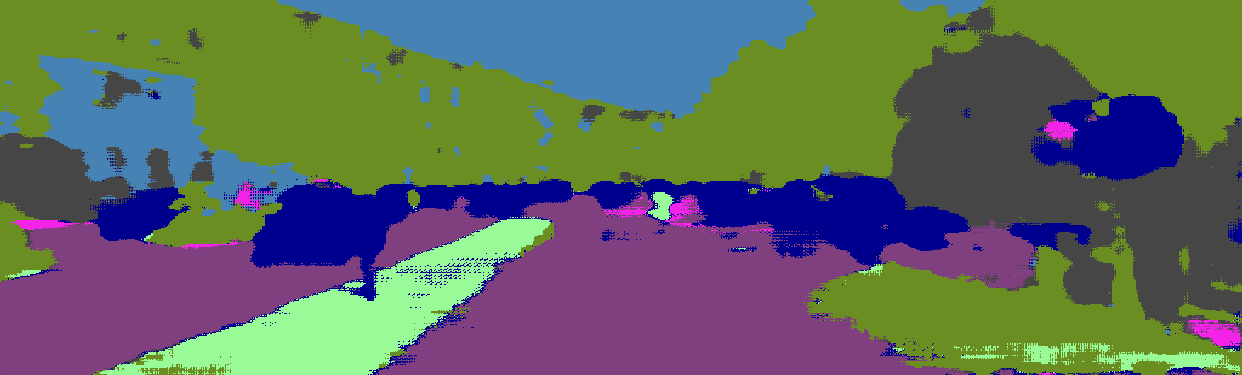}\vspace{0.1cm}
        \includegraphics[width=\linewidth]{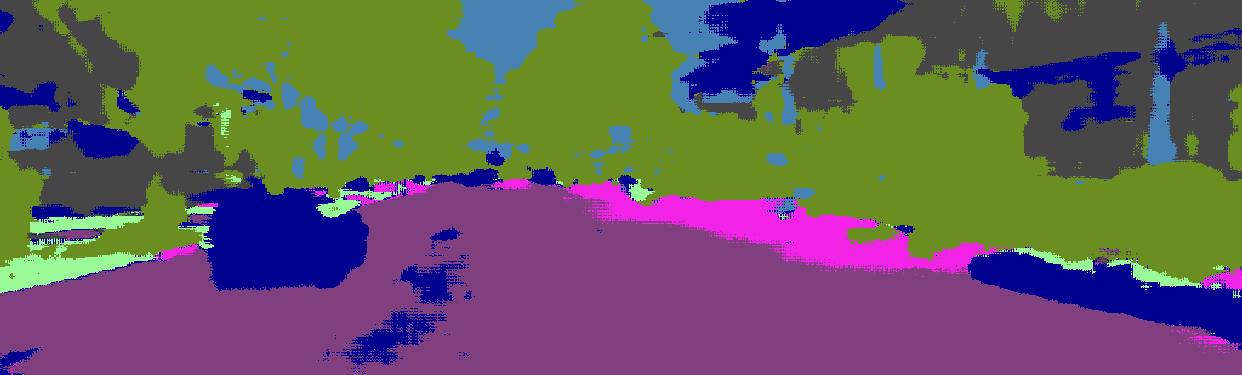}\vspace{0.1cm}
        \includegraphics[width=\linewidth]{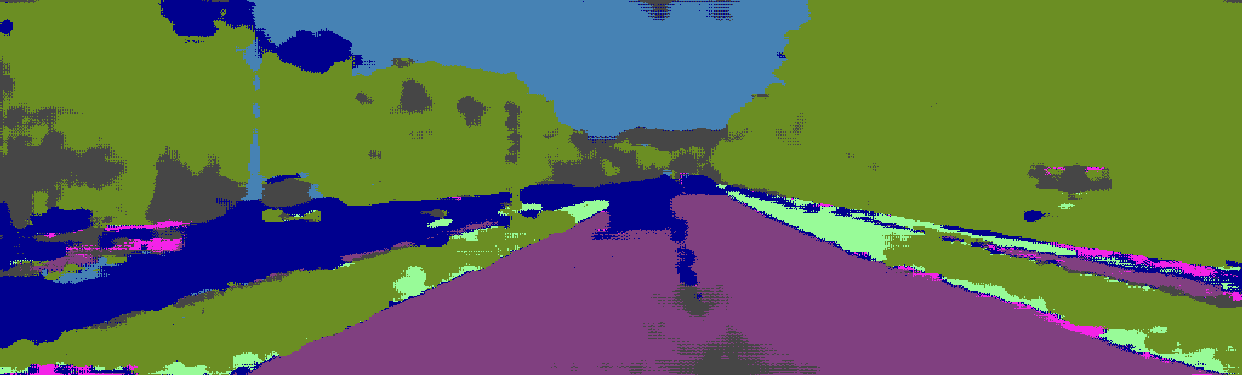}
    \end{minipage}\hfill
    \begin{minipage}{0.24\linewidth}
        \centering
        \caption*{Ours}
        \includegraphics[width=\linewidth]{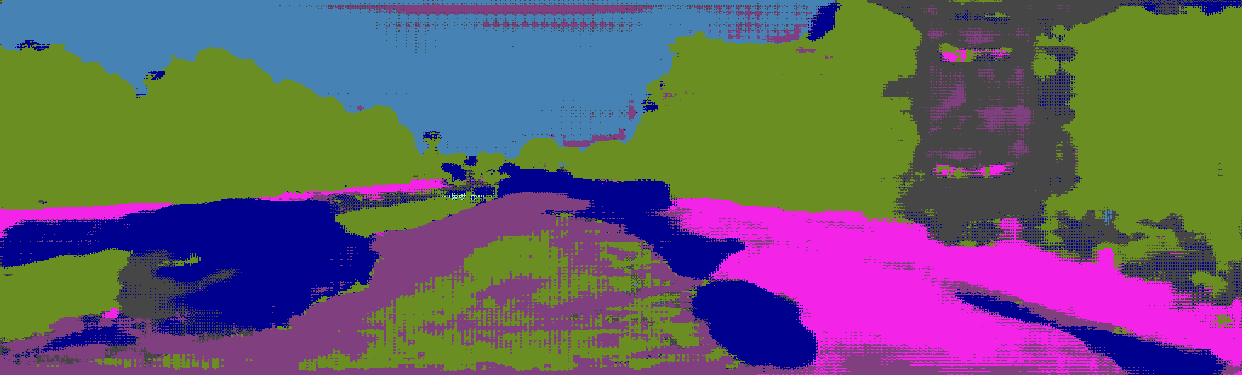}\vspace{0.1cm}
        \includegraphics[width=\linewidth]{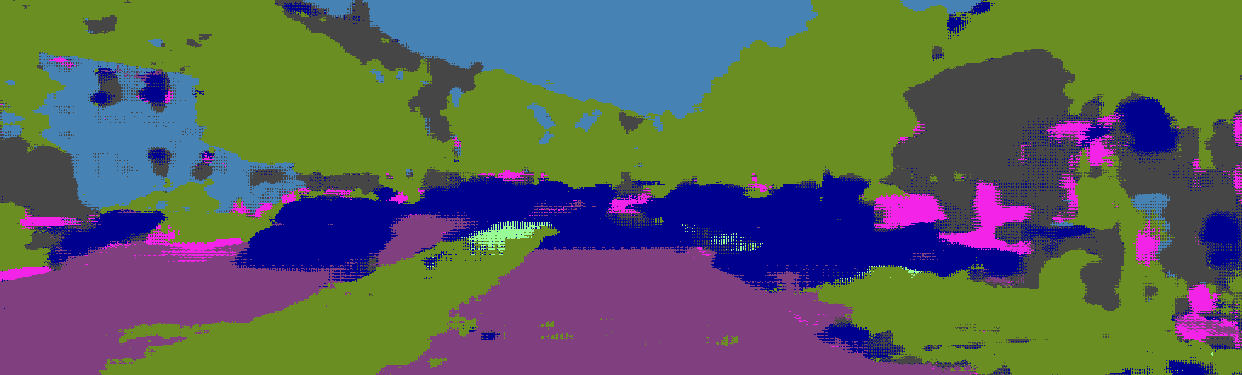}\vspace{0.1cm}
        \includegraphics[width=\linewidth]{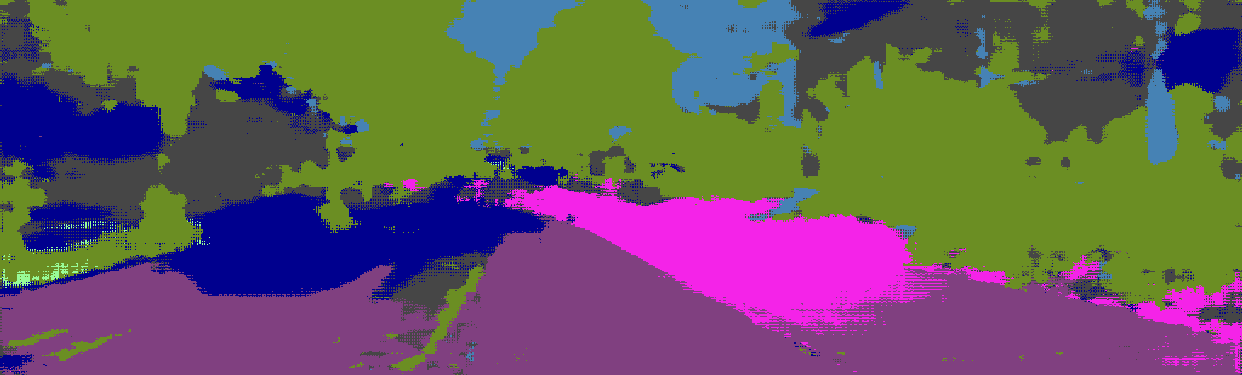}\vspace{0.1cm}
        \includegraphics[width=\linewidth]{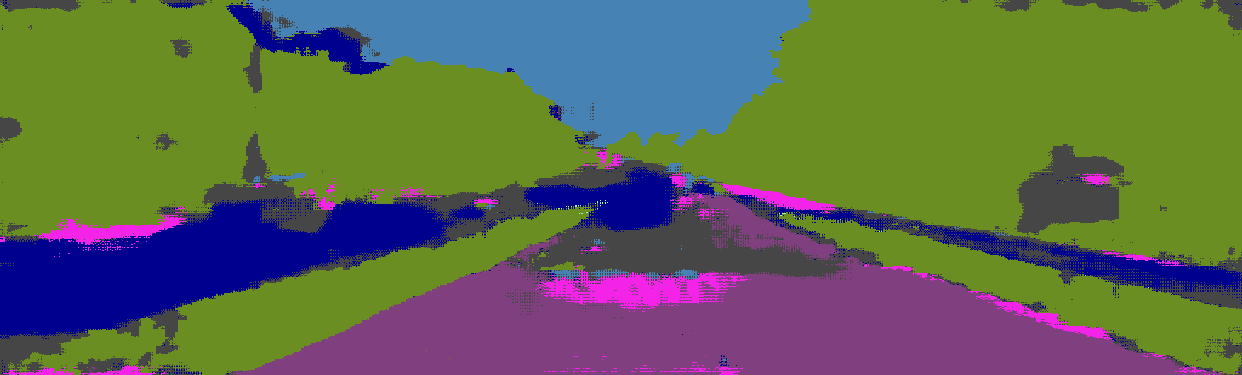}\vspace{0.1cm}
        \includegraphics[width=\linewidth]{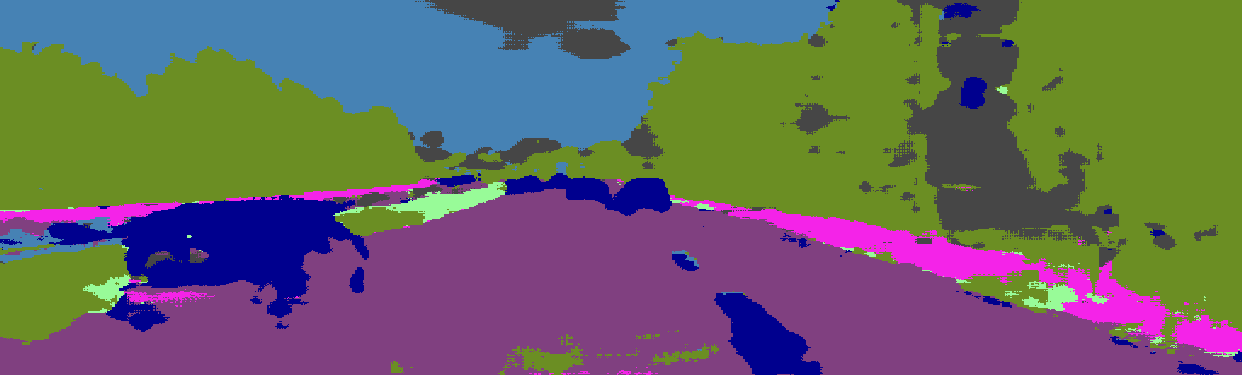}\vspace{0.1cm}
        \includegraphics[width=\linewidth]{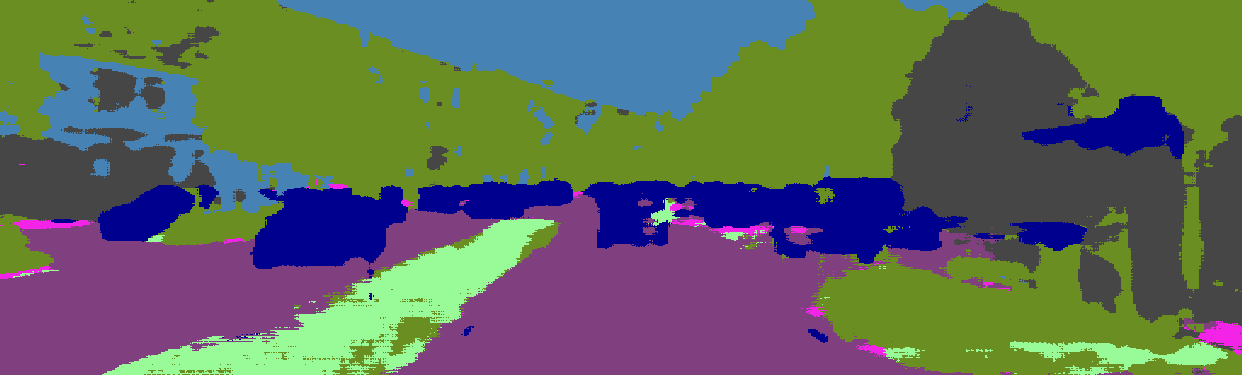}\vspace{0.1cm}
        \includegraphics[width=\linewidth]{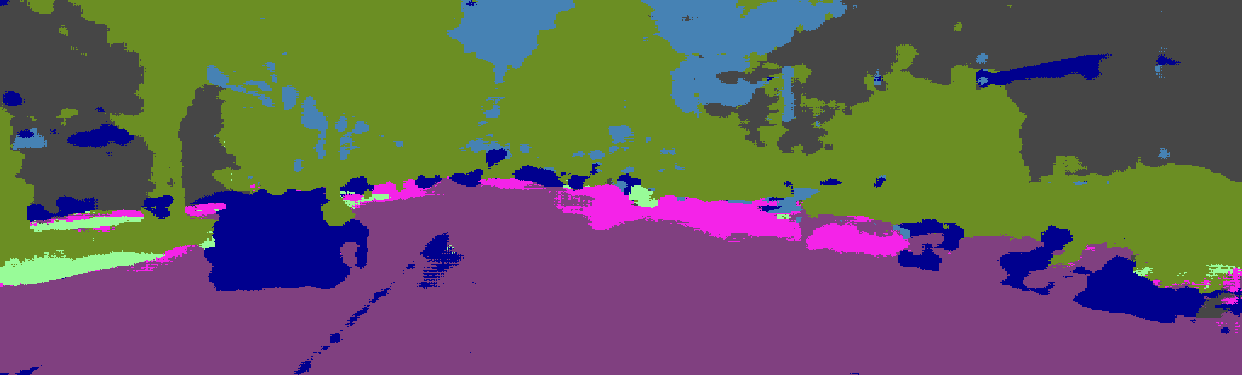}\vspace{0.1cm}
        \includegraphics[width=\linewidth]{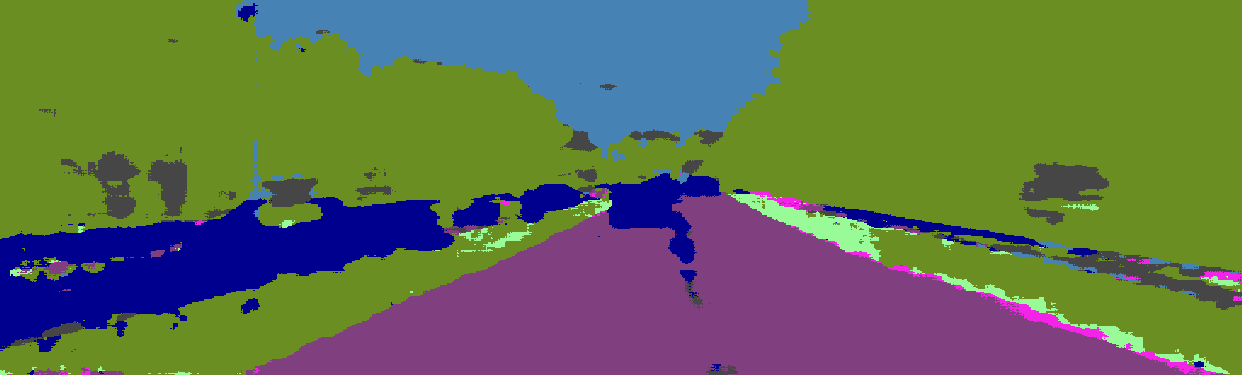}
    \end{minipage}
    \caption{Example results on the \textbf{KITTI2019} validation set (the four rows from the top are results from 1-shot class-incremental learning, and the last four rows are results from 5-shot learning).
    Left: Input image; Middle left: Ground-truth; Middle right: Prediction from the baseline method (FT+KD); Right: Prediction from our method (FT+KD+PL).}
    \label{img:kitti}
\end{figure*}

\noindent\textbf{Results on KITTI.} In Table~\ref{tab:kitti}, we present results on KITTI with Task 1, Task 2 and Task $1 \cup 2$ with 1-shot and 5-shot class-incremental learning, respectively. In most training stages, adding PL can significantly improve performance. For instance, with 5-shot class incremental learning, adding PL results in average mIoU improvements of 0.8\%, 4.5\%, and 0.8\% in Task~1, Task~2, and Task~$1 \cup 2$, respectively. We present some qualitative results in Figure~\ref{img:kitti}.

\section{Conclusion}
\label{sec:conclusion}

In this paper, we propose a simple and effective method for few-shot incremental semantic segmentation. Our method uses a scene level embedding network to retrieve unlabeled images with a similar scene layout to the support images, in order to perform effective pseudo-labeling and augment the few-shot training dataset. In addition, we employ knowledge distillation to prevent catastrophic forgetting of knowledge on previously learned classes. Experiments on two publicly available datasets in the self-driving domain validate the efficacy of the proposed approach.




\bibliographystyle{ieee}
\bibliography{sample}

\begin{thebibliography}{10}\itemsep=-1pt

\bibitem{abu2018augmented}
H.~Abu~Alhaija, S.~K. Mustikovela, L.~Mescheder, A.~Geiger, and C.~Rother.
\newblock Augmented reality meets computer vision: Efficient data generation
  for urban driving scenes.
\newblock {\em International Journal of Computer Vision}, 126:961--972, 2018.

\bibitem{cermelli2021prototype}
F.~Cermelli, M.~Mancini, Y.~Xian, Z.~Akata, and B.~Caputo.
\newblock Prototype-based incremental few-shot segmentation.
\newblock In {\em The 32nd British Machine Vision Conference}. BMVA Press,
  2021.

\bibitem{cordts2015cityscapes}
M.~Cordts, M.~Omran, S.~Ramos, T.~Scharw{\"a}chter, M.~Enzweiler, R.~Benenson,
  U.~Franke, S.~Roth, and B.~Schiele.
\newblock The cityscapes dataset.
\newblock In {\em CVPR Workshop on the Future of Datasets in Vision}, volume~2.
  sn, 2015.

\bibitem{deng2009imagenet}
J.~Deng, W.~Dong, R.~Socher, L.-J. Li, K.~Li, and L.~Fei-Fei.
\newblock Imagenet: A large-scale hierarchical image database.
\newblock In {\em 2009 IEEE conference on computer vision and pattern
  recognition}, pages 248--255. Ieee, 2009.

\bibitem{ganea2021incremental}
D.~A. Ganea, B.~Boom, and R.~Poppe.
\newblock Incremental few-shot instance segmentation.
\newblock In {\em Proceedings of the IEEE/CVF Conference on Computer Vision and
  Pattern Recognition}, pages 1185--1194, 2021.

\bibitem{he2016deep}
K.~He, X.~Zhang, S.~Ren, and J.~Sun.
\newblock Deep residual learning for image recognition.
\newblock In {\em Proceedings of the IEEE conference on computer vision and
  pattern recognition}, pages 770--778, 2016.

\bibitem{hinton2015distilling}
G.~Hinton, O.~Vinyals, and J.~Dean.
\newblock Distilling the knowledge in a neural network.
\newblock In {\em NIPS Workshops}, pages 1--9, 2014.

\bibitem{klingner2020class}
M.~Klingner, A.~B{\"a}r, P.~Donn, and T.~Fingscheidt.
\newblock Class-incremental learning for semantic segmentation re-using neither
  old data nor old labels.
\newblock In {\em ITSC}, pages 1--8, 2020.

\bibitem{li2017learning}
Z.~Li and D.~Hoiem.
\newblock Learning without forgetting.
\newblock {\em IEEE transactions on pattern analysis and machine intelligence},
  40(12):2935--2947, 2017.

\bibitem{lin2014microsoft}
T.-Y. Lin, M.~Maire, S.~Belongie, J.~Hays, P.~Perona, D.~Ramanan,
  P.~Doll{\'a}r, and C.~L. Zitnick.
\newblock Microsoft coco: Common objects in context.
\newblock In {\em Computer Vision--ECCV 2014: 13th European Conference, Zurich,
  Switzerland, September 6-12, 2014, Proceedings, Part V 13}, pages 740--755.
  Springer, 2014.

\bibitem{liu2020part}
Y.~Liu, X.~Zhang, S.~Zhang, and X.~He.
\newblock Part-aware prototype network for few-shot semantic segmentation.
\newblock In {\em Computer Vision--ECCV 2020: 16th European Conference,
  Glasgow, UK, August 23--28, 2020, Proceedings, Part IX 16}, pages 142--158.
  Springer, 2020.

\bibitem{michieli2021knowledge}
U.~Michieli and P.~Zanuttigh.
\newblock Knowledge distillation for incremental learning in semantic
  segmentation.
\newblock {\em Computer Vision and Image Understanding}, 205:103167, 2021.

\bibitem{nguyen2022ifs}
K.~Nguyen and S.~Todorovic.
\newblock ifs-rcnn: An incremental few-shot instance segmenter.
\newblock In {\em Proceedings of the IEEE/CVF Conference on Computer Vision and
  Pattern Recognition}, pages 7010--7019, 2022.

\bibitem{shi2022incremental}
G.~Shi, Y.~Wu, J.~Liu, S.~Wan, W.~Wang, and T.~Lu.
\newblock Incremental few-shot semantic segmentation via embedding
  adaptive-update and hyper-class representation.
\newblock In {\em Proceedings of the 30th ACM International Conference on
  Multimedia}, pages 5547--5556, 2022.

\bibitem{su2020does}
J.-C. Su, S.~Maji, and B.~Hariharan.
\newblock When does self-supervision improve few-shot learning?
\newblock In {\em European conference on computer vision}, pages 645--666.
  Springer, 2020.

\bibitem{tao2020few}
X.~Tao, X.~Hong, X.~Chang, S.~Dong, X.~Wei, and Y.~Gong.
\newblock Few-shot class-incremental learning.
\newblock In {\em Proceedings of the IEEE/CVF Conference on Computer Vision and
  Pattern Recognition}, pages 12183--12192, 2020.

\bibitem{wang2017efficient}
T.~Wang, X.~He, S.~Su, and Y.~Guan.
\newblock Efficient scene layout aware object detection for traffic
  surveillance.
\newblock In {\em Proceedings of the IEEE Conference on Computer Vision and
  Pattern Recognition Workshops}, pages 53--60, 2017.

\end{thebibliography}

\end{document}